\documentclass[pdflatex,sn-vancouver-num]{sn-jnl}

\usepackage{graphicx}%
\usepackage{multirow}%
\usepackage{amsmath,amssymb,amsfonts}%
\usepackage{amsthm}%
\usepackage{mathrsfs}%
\usepackage[title]{appendix}%
\usepackage{xcolor}%
\usepackage{textcomp}%
\usepackage{manyfoot}%
\usepackage{booktabs}%
\usepackage{array}
\usepackage{ragged2e}
\usepackage{changepage}

\begin{document}

\title[Quantized and Efficiently Adapted Protein Language Models]{Analysis of Quantized and Efficiently Adapted Protein Language Models}

\author[1]{\fnm{Ilan Yaniv} \sur{Zeisler}}
\equalcont{These authors contributed equally to this work.}

\author[1]{\fnm{Sebastian} \sur{Clancy}}
\equalcont{These authors contributed equally to this work.}

\author[1]{\fnm{Pouriya} \sur{Bayat}}

\author[4]{\fnm{Saaim} \sur{Raad}}

\author[3]{\fnm{Ivan} \sur{Kraskov}}

\author[3]{\fnm{Matthew} \sur{Xie}}

\author[1]{\fnm{Vivian} \sur{White}}

\author[2]{\fnm{Spencer} \sur{Perkins}}

\author[1]{\fnm{Serena} \sur{Singh}}

\author[5]{\fnm{Sepehr} \sur{Bayat}}

\author*[1,2]{\fnm{Keith} \sur{Pardee}}\email{keith.pardee@utoronto.ca}

\affil[1]{\orgdiv{Leslie Dan Faculty of Pharmacy}, \orgname{University of Toronto}, \orgaddress{\city{Toronto}, \state{ON}, \country{Canada}}}
\affil[2]{\orgdiv{Department of Mechanical and Industrial Engineering}, \orgname{University of Toronto}, \orgaddress{\city{Toronto}, \state{Ontario}, \country{Canada}}}
\affil[3]{\orgdiv{Department of Computer Science}, \orgname{University of Toronto}, \orgaddress{\city{Toronto}, \state{Ontario}, \country{Canada}}}
\affil[4]{\orgdiv{Department of Electrical and Computer Engineering}, \orgname{University of Toronto}, \orgaddress{\city{Toronto}, \state{Ontario}, \country{Canada}}}
\affil[5]{\orgdiv{Department of Computing and Software}, \orgname{McMaster University}, \orgaddress{\city{Hamilton}, \state{ON}, \country{Canada}}}

\abstract{\textbf{Background:} Protein language models (PLMs) are increasingly used for sequence generation and protein property prediction, but their growing size makes fine-tuning and deployment computationally expensive. Quantization and parameter efficient fine-tuning (PEFT) may reduce these requirements, but their effects on PLM performance, representations, and generative behavior are insufficiently characterized.

    \textbf{Results:} We systematically evaluated 4-bit quantization and low-rank adapter fine-tuning (QLoRA) across gold standard PLM architectures and scales, including ESM-2, ESMC, ProtBERT, ProtT5, Ankh, Ankh3 and Profluent-E1. Across diverse protein prediction tasks, QLoRA broadly retained downstream performance compared with full fine-tuning, with many model-task pairs retaining more than 90\% of full-model performance while substantially reducing peak GPU memory usage. The largest models showed the greatest memory savings, with reductions approaching 90\%, although performance and efficiency varied by model family, dataset and training configuration. Layer-wise representation analysis showed that QLoRA often preserved early-layer representations while inducing task-specific adaptations in middle and late layers, broadly resembling full fine-tuning but with lower-magnitude representational change. Computational benchmarks showed substantial reductions in peak GPU memory usage, with more varied effects on training speed and power usage. We also evaluated 4-bit quantization for unconditional sequence generation with ProLLaMA, ProtGPT2, ProGen2, ProteinGLM and ESM3. Quantized generative models largely preserved predicted structural and sequence-level properties as compared to full models, but token-level analysis revealed model-dependent shifts in autoregressive output distributions.

    \textbf{Conclusion:} QLoRA and 4-bit quantization provide practical methods for reducing the computational requirements of PLM fine-tuning and unconditional generation, particularly GPU memory usage. For supervised prediction tasks, our results support a workflow in which QLoRA is used as a first-pass strategy for memory limited PLM adaptation, with full fine-tuning reserved for challenging tasks, unstable architectures or low validation recovery. For generative PLMs, standard sequence-level and structural metrics should be complemented with distributional analysis since quantization-induced shifts in autoregressive output distributions may not be apparent from downstream predictions alone. These findings support the use of quantization and efficient fine-tuning methods to broaden access to large-scale protein modelling while emphasizing the need for model- and task-specific validation.
}

\keywords{Proteins, Machine Learning, Protein Language Models, Quantization, Parameter-Efficient Fine-Tuning, Unconditional Generation}

\maketitle

\newpage

\section{Background}\label{sec:background}

\begin{figure}[htbp]
    \centering
    \includegraphics[width=\textwidth]{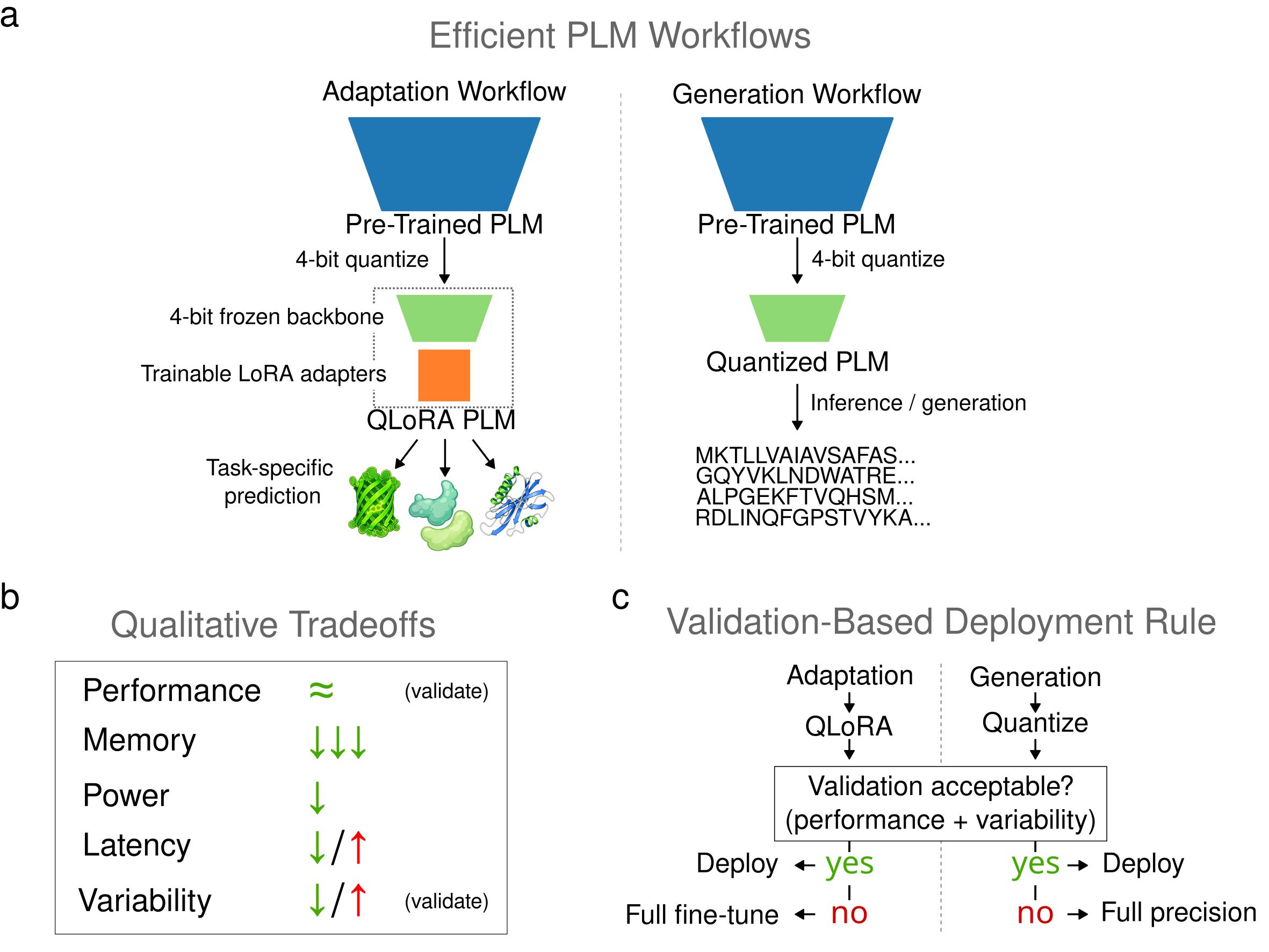}
    \caption{
        Efficient use of pretrained protein language models for both property prediction and protein generation.
        \textbf{a} We evaluated PLMs using two different tasks. For fine-tuning (adaptation) to a specific dataset (left), the pretrained PLM backbone is quantized to 4-bit precision and augmented with trainable LoRA adapters,
        producing a QLoRA model. For protein generation (right), the PLM is quantized without added LoRA adapters.
        \textbf{b} To better understand the effects of quantization, we looked at several metrics for success, including performance (accuracy), training memory, power consumption, latency, and variability.
        Model performance was maintained, while memory and power improvements were observed.
        \textbf{c} The suggested practitioner workflow derived from benchmarking results. QLoRA is used as
        a first-pass strategy for memory-constrained supervised adaptation, with fallback to full fine-tuning for difficult tasks, unstable
        architectures or low-performance recovery. For quantized generation, aggregate sequence and structural metrics should be supplemented
        with distributional diagnostics to evaluate downstream use cases, falling back to the full precision model in some cases.
    }
    \label{fig:overview}
\end{figure}

Machine learning (ML) and protein language models (PLMs) in particular have become central tools for protein sequence design, analysis, structure prediction and functional annotation \cite{zhao2023survey,Vieira2024.11.22.624936}. Like large language models more broadly, PLMs have grown increasingly larger and benefited from more training data and increased compute \cite{kaplan2020,zhao2023survey,lin2023evolutionary}. However, this has also resulted in heavier computational resource requirements for model training, fine-tuning and inference, a limitation that is particularly acute for many academic laboratories and smaller organizations \cite{han2024parameterefficientfine-tuninglargemodels}. Methods for reducing the cost of fine-tuning and deploying existing PLMs are thus important for making these models more accessible \cite{han2024parameterefficientfine-tuninglargemodels}.

Scaling laws have found that for LLMs (and PLMs), most model improvements and architectural choices have minimal impact on performance when compared to scaling up the compute used and training data \cite{kaplan2020}. This has led to models that reach trillions of parameters, and PLMs have followed this trend as well \cite{zhao2023survey, Vieira2024.11.22.624936}. Current methods for reducing the computational cost of large models focus either on creating new architectures that use fewer parameters, or modifying existing training and inference methods to reduce model size \cite{han2024parameterefficientfine-tuninglargemodels}. We focus on methods that integrate with existing models, in particular quantization and parameter efficient fine-tuning methods.

Quantization, in terms of machine learning, represents model parameters using lower-precision numerical formats,
reducing memory requirements compared to standard 16- or 32-bit formats, but often at a cost to accuracy \cite{gholami2021survey,jacob2018quantization}.
In this study, we focus on 4-bit quantization which substantially reduces the memory needed to store model parameters \cite{dettmers2024qlora}.
However, aggressive quantization can alter model outputs or reduce downstream task performance, especially when models are adapted to new tasks \cite{gholami2021survey}.

Low Rank Adaptation (LoRA) is a parameter-efficient fine-tuning (PEFT) method that freezes pretrained model weights and trains small low-rank adapter matrices instead of all parameters in the model \cite{lora}.
Along with other PEFT methods, LoRA has been used in LLMs to greatly reduce their trainable parameters, making it easier to fine-tune \cite{lora}.
Dettmers et al. (2024) showed that combining quantization of base model weights with LoRA (QLoRA) enables efficient fine-tuning of LLMs with reduced memory requirements \cite{dettmers2024qlora}.
LoRA has been applied to PLMs for tasks such as protein-protein interaction, protein property prediction and structural
classification \cite{LoraNatureComm, berger-et-al-2024}, and quantization has been applied to PLMs both on its own, in the
post-training setting \cite{peng2023ptq}, and alongside parameter-efficient fine-tuning of billion-parameter
models \cite{celik2025efficient,LoraNatureComm}.
The extent to which QLoRA preserves performance and internal model representations
across diverse PLM architectures and sizes, however, remains less well characterized, and to the best of our knowledge has not been
evaluated systematically across model families and scales.

Quantization can also affect generative PLMs independently of fine-tuning \cite{gong2024quantization,peng2023ptq}.
Autoregressive and iterative masked decoding models are increasingly used for protein sequence generation, but changes in weight precision could change
the distribution of generated protein sequences and their properties \cite{ferruz2022protgpt2,nijkamp2023progen2,hayes2025esm3,lv2024prollama}.

In this study (Fig.~\ref{fig:overview}), we systematically evaluate QLoRA fine-tuning across multiple PLM families and scales, comparing QLoRA
with full fine-tuning for various protein prediction tasks and assess layer-wise representation differences. We further examine the effects of
4-bit quantization when creating random sequences (unconditional generation), in order to evaluate whether quantization preserves the model's
learned distribution. Lastly, we benchmark memory, training throughput and power usage for fine-tuning and generation. We found that applying QLoRA to
PLMs reduced peak GPU memory during fine-tuning by an average of 63\%, approaching 90\% for the largest models, while many model-task pairs retained
more than 90\% of full fine-tuning performance. Together, these analyses
evaluate whether quantization and PEFT can reduce the computational cost of PLMs while preserving utility and democratizing access.

\section{Methods}\label{sec:methods}

\subsection{Pretrained PLMs}\label{subsec:plms}

All models are based on the transformer architecture and initialized with pretrained checkpoints from the Hugging Face Transformers repository.
For fine-tuning with QLoRA, we included ProtBERT (420M parameters) \cite{elnaggar2022prottrans}, ProtT5-half (3B) \cite{elnaggar2022prottrans}, Ankh-Base (726M) \cite{elnaggar2023ankh}, Ankh3-Large (1.9B) \cite{alsamkary2025ankh3} and several model families with different parameter
counts including ESM-2 (8M, 35M, 150M, 650M, 3B) \cite{lin2023evolutionary}, ESMC (300M, 600M) \cite{esmteam2024esmc} and Profluent-E1 (150M, 300M, 600M) \cite{jain2025e1}.
For unconditional generation with quantization, we used ProtGPT2 (738M) \cite{ferruz2022protgpt2}, ProLLaMA (6.7B) \cite{lv2024prollama}, ESM3 (1.4B) \cite{hayes2025esm3}, ProGen2 XL (6B) \cite{nijkamp2023progen2} and ProteinGLM (7B) \cite{chen2025xtrimopglm}.

\subsection{QLoRA Fine-tuning}\label{subsec:qlora-ft}

We used the following datasets for fine-tuning: GFP brightness \cite{sarkisyan2016local}, protease stability \cite{rocklin2017}, protein secondary structure \cite{pdb, casp, netsurfp}, protein solubility \cite{torchdrug}, GB1 binding \cite{gb1_bioproject}, and PolyA binding \cite{melamed2013polya}. We performed an initial hyperparameter search on ESM-2 150M across learning rate (LR), alpha, and rank, and used this as a basis for fine-tuning all the models with adjustments to LR and batch size to facilitate convergence as needed (Table~S2). For QLoRA models, we quantized to 4-bit float \cite{dettmers2024qlora} on linear layers in the base model (either by default layers specified in the model implementation in the Transformers library or our specification if there was no default) and used LoRA rank=8, alpha=32 and dropout=0.05. We fine-tuned each model on each dataset task for both the full model and the QLoRA model in triplicate across three seeds for 20 epochs, with early stopping enabled using a patience of three. We evaluated models with Spearman's rank correlation coefficient (SpearmanR) for each task except for the secondary structure classification task, in which we used multi-class balanced accuracy.

\subsection{Fine-tuning Representation Analysis}\label{subsec:rep-analysis}

The effects of QLoRA fine-tuning on internal representations of PLMs were assessed by comparing layer-wise representations across pretrained (without fine-tuning), fully fine-tuned and QLoRA fine-tuned variants using held-out test splits for each downstream task and in triplicate with three random seeds. We considered three pairwise comparisons for each architecture: pretrained versus fully fine-tuned, pretrained versus QLoRA fine-tuned, and fully fine-tuned versus QLoRA fine-tuned, allowing us to quantify both the extent to which fine-tuning altered the pretrained representation space and whether QLoRA substantially differed in representational geometry as compared to full fine-tuning.

For each variant, hidden states were extracted from all transformer layers with models in evaluation mode using the same input sequences, tokenization and attention masks between each pair. Padding positions were excluded from all token-level analysis. To facilitate comparisons across architectures of different depth, layer depths were normalized with first layers assigned depth 0 and final layers assigned depth 1. Representation similarity was quantified using centered kernel alignment (CKA), wherein token-level hidden states for each layer across variant pairs were mean-centered and compared as

\begin{equation}
    \mathrm{CKA}(X, Y) = \frac{\|Y^\top X\|_F^2}{\|X^\top X\|_F \; \|Y^\top Y\|_F}
\end{equation}

where X and Y denote corresponding hidden-state matrices for the two models. CKA was computed at each layer independently, providing a depth-resolved profile of representation similarity for each model, task and seed. Hidden states were projected to a common lower-dimensional space using fixed random projections before computing CKA to make analysis tractable for large PLMs.

We further evaluated the task relevance of each layer using a non-parametric k-nearest-neighbor (kNN) probe. For each layer, hidden states were mean-pooled over non-padded positions to obtain a single sequence-level embedding per protein. A leave-one-out kNN predictor was then applied with cosine similarity and k = 10. For regression tasks, each prediction was computed as a mean label value among nearest neighbors and evaluated using Spearman rank correlation. For secondary-structure classification, we used a sequence-level proxy task by assigning each sequence the majority residue-level class and evaluated predictions with macro-averaged F1 score. Since no probe parameters were optimized, the analysis provides a low-capacity, non-parametric estimate of label-relevant neighborhood structure in frozen representation space.

\subsection{Quantized Unconditional Generation}\label{subsec:uncond-gen}

To evaluate the effects of 4-bit quantization on unconditionally generated sequences and their predicted biological and structural properties, we compared unconditional samples from full-precision and 4-bit quantized variants of each generative PLM. Analyses were performed on ProLLaMA, ProtGPT2, ProGen2, ProteinGLM and ESM3, with sequences generated without conditioning on input sequence or annotation. Autoregressive models were sampled using their model-specific generation procedures while ESM3 was sampled using iterative generation.

For each model variant and random seed, 250 sequences were generated for each length bin of 100-150, 150-200 and 200-300 residues. For each retained sequence, we recorded sequence length and evaluated structural plausibility, structural similarity to known proteins, predicted secondary structure and predicted disorder. Predicted structures were generated using Boltz-2 \cite{passaro2025boltz2} with a multiple-sequence alignment constructed locally using ColabFold 1.5 and MMseqs2.18 with UniRef30 and ColabFoldDB databases. From the Boltz-2 outputs, we extracted the mean predicted local distance difference test score (pLDDT) across residues and the predicted template modelling (pTM) score for each sequence. These metrics represent model-based estimates of local structure confidence and global fold plausibility, respectively. We used Foldseek \cite{van2024fast} (release 10 with PDB database) to assess similarity to known protein structures in the Protein Data Bank. For each sequence, the best structural hit was summarized using Foldseek-reported LDDT, alignment TM-score and homology probability. We also recorded the Foldseek pass rate, defined as the fraction of sequences with any homology match.

We used S4Pred \cite{moffat2021increasing} 1.2 to determine helix, sheet and coil content with low-confidence predictions filtered using the S4Pred confidence threshold before class fractions were calculated. Intrinsic disorder was predicted using AIUPred \cite{erdHos2024aiupred} 1.2, providing per-residue disorder probabilities for each sequence that were used to calculate a mean sequence-level disorder score.

\subsection{Teacher Forced KL Divergence}\label{subsec:kl}

To quantify the effect of 4-bit quantization on distributions of autoregressive PLMs, we computed token-level Kullback-Leibler (KL) divergence between full-precision and quantized model variants under teacher forcing. Since this analysis is restricted to causal language models defining a left-to-right next-token likelihood, we analyzed ProLLaMA, ProtGPT2, ProGen2 and ProteinGLM. ESM 3 was excluded because its sampling procedure does not provide an equivalent autoregressive next-token factorization.

Evaluation was performed using sequences sampled from the Swiss-Prot subset of UniProt and filtered to contain only canonical amino acids and lengths between 100 and 300 residues. Sequences were further stratified into three length bins: 100-150, 150-200 and 200-300 residues. For each bin, a fixed number of sequences were drawn with a fixed random seed, providing a held-out evaluation set independent of model generation.

For each sequence, full-precision and quantized models were evaluated on identical ground-truth prefix contexts. Models were not allowed to autoregressively sample; instead teacher-forcing was used in order to isolate differences in conditional next-token distributions from differences caused by error accumulation or stochastic decoding during generation. For each position i, a model was provided the true sequence of tokens up to that position and produced a probability distribution over the next token. Let $P_{\mathrm{full}}(\cdot \mid x_{\le i})$ and $P_{\mathrm{quant}}(\cdot \mid x_{\le i})$ denote full-precision and quantized next-token distributions for identical prefix context. For each token position, we computed the KL divergence in the direction of full-precision model to quantized model, treating the full-precision distribution as reference:

\begin{equation}
    D_{\mathrm{KL}}(P_{\mathrm{full}} \,\|\, P_{\mathrm{quant}})
    =
    \sum_v P_{\mathrm{full}}(v \mid x_{\le i})
    \Big[
        \log P_{\mathrm{full}}(v \mid x_{\le i})
        -
        \log P_{\mathrm{quant}}(v \mid x_{\le i})
        \Big]
\end{equation}

where the sum is over the model vocabulary. We also computed entropy of the full-precision distribution for each position to contextualize divergence relative to the reference model's intrinsic predictive uncertainty:

\begin{equation}
    H(P_{\mathrm{full}}) = -\sum_v P_{\mathrm{full}}(v \mid x_{\le i}) \log P_{\mathrm{full}}(v \mid x_{\le i})
\end{equation}

Token-level KL divergence and entropy values were retained for each model variant, length bin, sequence, position and seed. Sequence positions were further represented as normalized positions in [0,1] to enable comparisons across different lengths. Summary analyses aggregated per-token divergence across seeds and length bins, and divergence was analyzed as a function of reference entropy, sequence length and normalized sequence position.

\subsection{Compute Benchmark}\label{subsec:compute}

We evaluated several metrics of computational cost for fine-tuning each model. We tracked the maximum GPU memory utilized during training for each model, the average power consumption of the graphics card during training, and also the training speed in steps per second. In each case, we trained the model for various batch sizes (1, 2, 4,..., 28) and measured the metrics for both the full and QLoRA models. To account for variation in the training process, each set of evaluations was repeated across three different seeds. Each metric was calculated as the percent difference of the quantized to full model, as well as standard deviation for that ratio. We allowed the model to train for 3 warmup steps before we began recording these metrics for a window of 15 training steps. These models were trained on the GFP mutant fluorescence prediction task.

For evaluating unconditional sequence generation, we generated 10 sequences after a 2 sequence warmup with a maximum output length of 200 tokens per sequence, and this was repeated across three random seeds for each model. We calculated the quantized to full model ratio of amino acids generated per second, peak GPU memory allocation and average GPU power draw.

\subsection{Hardware \& Software}\label{subsec:hardware}

Model fine-tuning was performed on NVIDIA GeForce RTX 4090s (Ada, sm\_89), 5090s (Blackwell, sm\_120) and a DGX Spark (GB10 Grace Blackwell, sm\_121), depending on availability. Representation analyses and unconditional generation benchmarks were performed on RTX 5090s. Unconditional generation and sequence analysis were performed on the DGX Spark. All experiments were performed using Python 3.12 with PyTorch 2.11+cu128, Hugging Face Transformers 4.48, PEFT 0.18, bitsandbytes 0.49, and Accelerate 1.13.

\section{Results}\label{sec:results}

\subsection{Fine-Tuning Performance}\label{subsec:ft-perf}

\begin{figure}[htbp]
    \centering
    \includegraphics[width=\textwidth]{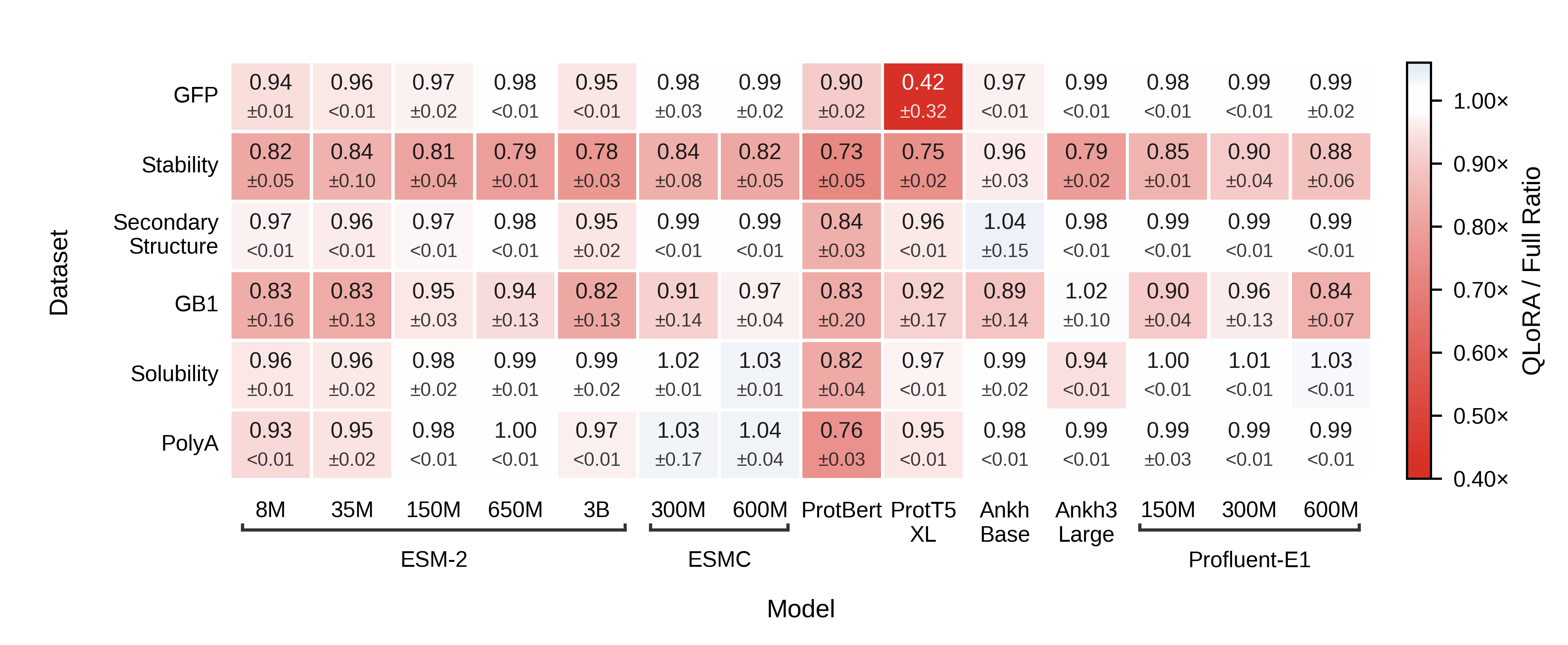}
    \caption{
        QLoRA fine-tuning performance ratios across models and tasks.
        Each value shows the relative performance (full model score / QLoRA model score) in large text, with standard deviation across three replicates ($n=3$ seeds) shown below.
    }\label{fig:ft-ratios}
\end{figure}

With the goal of evaluating QLoRA performance in comparison to full precision models, we fine-tuned each model on several datasets.
Each model was trained to predict a specific property or aspect of protein sequences, and then evaluated in triplicate with different training seeds to account for training instability.
We then compared the results to the original models, seeing that QLoRA generally performed similarly ($\pm5\%$) to the full model (Fig.~\ref{fig:ft-ratios}).
For smaller models, batch size and learning rate could be kept the same for full and QLoRA models to obtain convergence, while larger models tended to
require a magnitude lower learning rate for the full model relative to the QLoRA model (Table~S2).
ESMC 600M, the larger of the two latest generation ESM models, performed best in both overall fine-tuning performance and in retaining QLoRA performance relative
to the full fine-tuned model (Fig.~S1).

\begin{figure}[htbp]
    \centering
    \includegraphics[width=\textwidth]{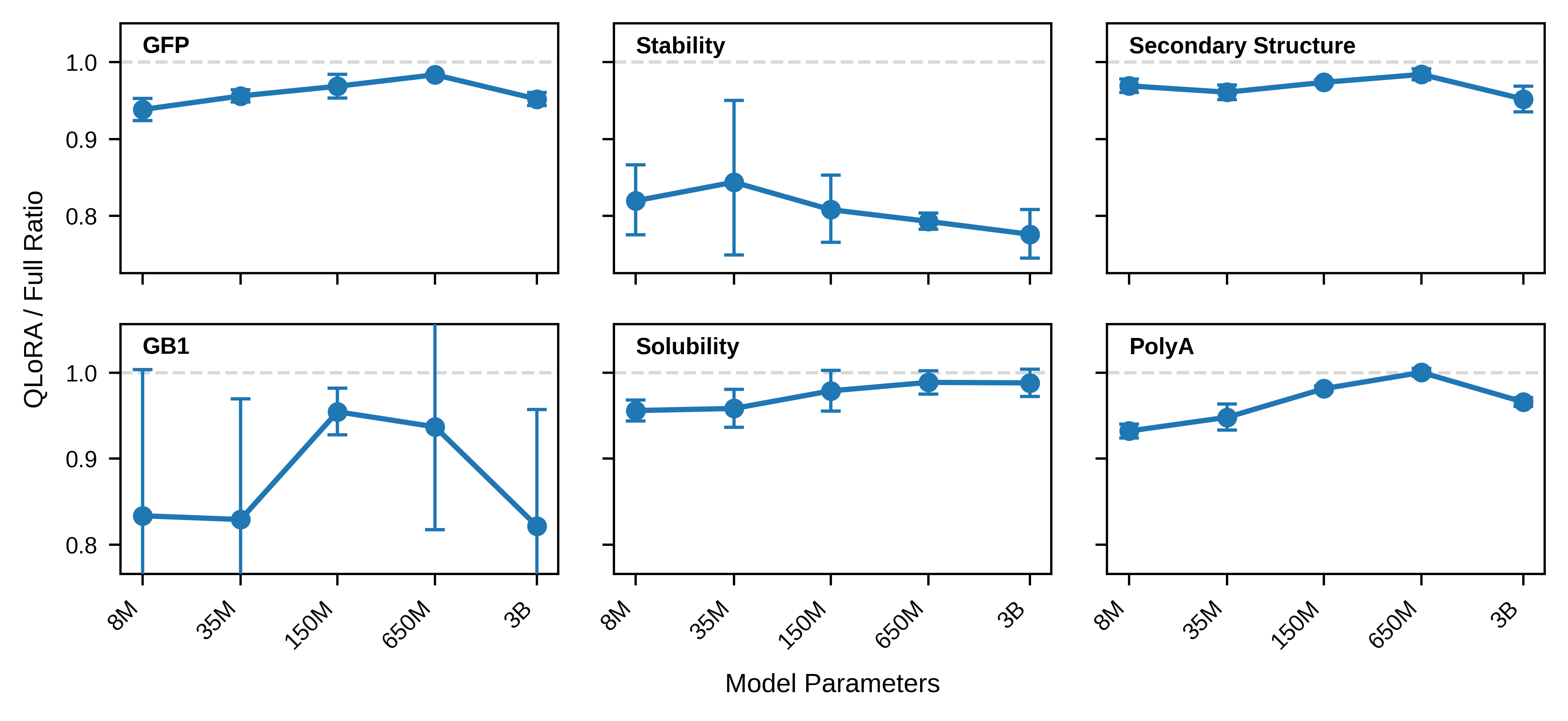}
    \caption{Relative QLoRA/full fine-tuning performance of ESM-2 models for different tasks by model size. Ratios are geometric means with 95\% confidence interval error bars across three replicates ($n=3$ seeds).}\label{fig:ft-size}
\end{figure}

There were some general trends that we observed in our analysis of PLM performance. Firstly, both full and QLoRA models had generally poor performance on the
stability, solubility and GB1 datasets (Fig.~S1) with pronounced variance across all models for stability
and GB1 (Fig.~\ref{fig:ft-size}). This has been reported before: these tasks pair small labelled datasets with a narrow dynamic range of measured values, and
fully fine-tuned PLMs also achieve comparatively low correlation on them, so the weak recovery seen here reflects the difficulty of the task rather than an
effect of quantization \cite{dallago2021flip,LoraNatureComm}.
Interestingly, for the ESM-2 family of models (Fig.~\ref{fig:ft-size}), which have the widest number and range of size variants, better relative QLoRA
performance was typically observed as parameterization increased from 8M to 650M, followed by worse relative performance often at 3B.
ESMC 300M and 600M showed performance similar or better than the much larger
ESM-2 3B. The T5 based models, ProtT5 and Ankh, were both challenging to train and often exhibited high sensitivity to learning rate and batch size with large
variance (Fig.~\ref{fig:ft-size}; Fig.~S1).

\subsection{Representation Analysis}\label{subsec:rep-results}

\begin{figure}[htbp]
    \centering
    \includegraphics[width=\textwidth]{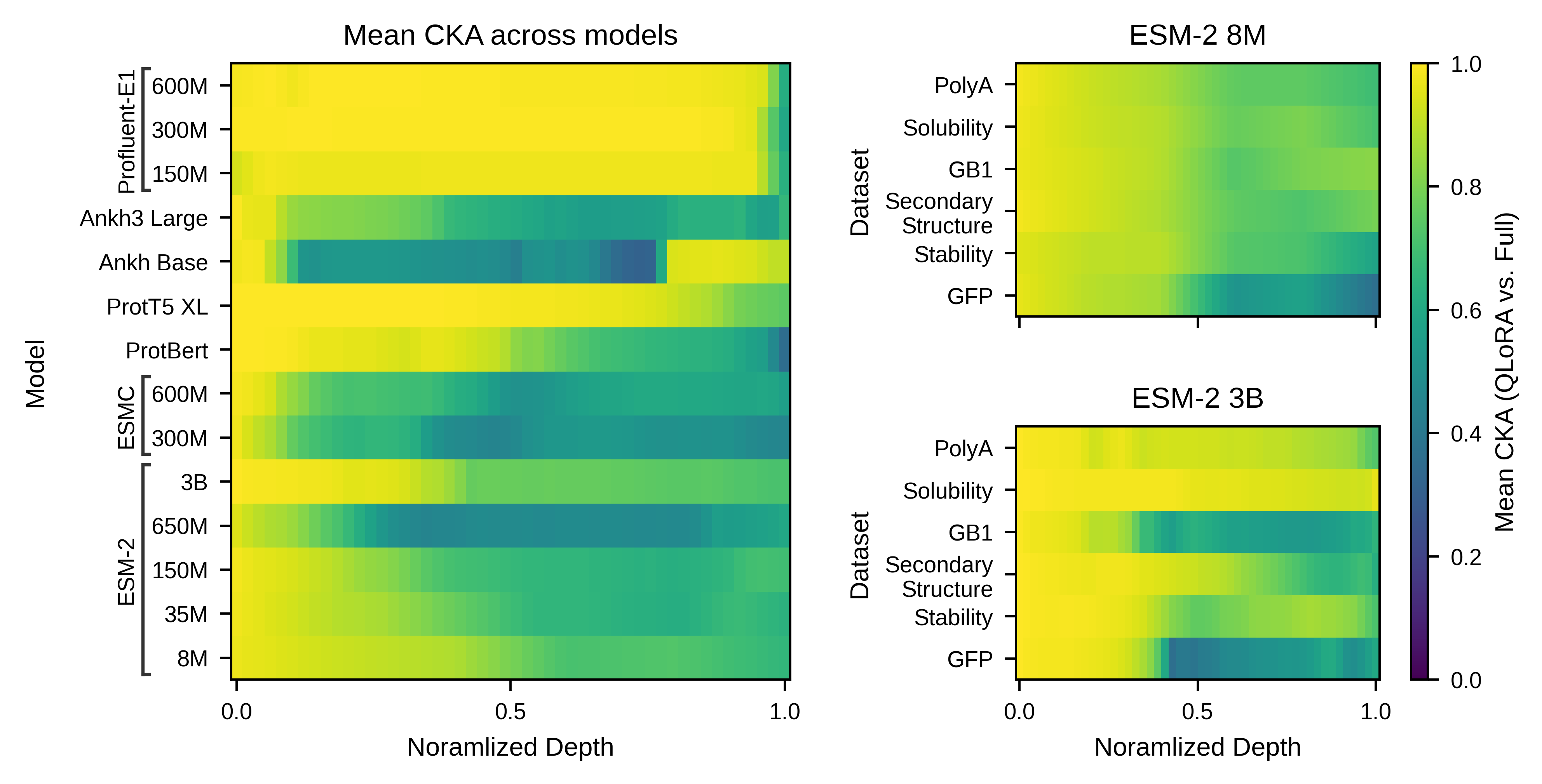}
    \caption{Layer-wise representational similarity between QLoRA and full fine-tuning, with higher mean CKA indicating higher similarity to the full model. \textbf{a} Mean CKA similarity averaged across datasets and three paired seeds for each model as a function of normalized layer depth. \textbf{b} Dataset-specific CKA profiles for ESM-2 8M and 3B, averaged across three paired seeds.}\label{fig:cka}
\end{figure}

While task performance comparisons provide an external measure of fine-tuning quality, they offer limited insight into how QLoRA and full fine-tuning reshape
internal representations in PLM layers. In order to examine these differences, here we analyze representation structure using centered kernel alignment (CKA) and kNN task probes, which provide
insight into how closely adapted representations remain aligned with the pretrained model and how each layer progressively organizes proteins according
to task-relevant properties. CKA analysis shows QLoRA fine-tuning often preserves early-layer representation in the PLM neural network relative to full fine-tuning
(yellow, Fig.~\ref{fig:cka}a), but diverges in the middle and late layers depending on the model (green/blue). Nearly all models show very high CKA near the
beginning layers suggesting that QLoRA (Fig.~S2b) and full-fine-tuning (Fig.~S2a) leave low-level sequence representations mostly aligned. Large differences appear in intermediate and later layers where
task-dependent adaptation is expected to be largest. This is consistent with findings about LLM layer importance, where studies have found that these intermediate and
later layers encode task specific information \cite{jin2024conceptdepth,tenney2019bert,jawahar2019bert}.

The Profluent-E1 model family shows the strongest similarity between QLoRA and full model fine-tuning with CKA remaining close to 1.0 (unchanged) for
the majority of the depth range and only a modest drop at the final layers. This suggests that QLoRA retains a representation
geometry very similar to that of full fine-tuning. ProtT5 similarly preserves most representation between the full and QLoRA
variants across most layers of the neural network, with divergence concentrated near the final layers, consistent with adaptation being focused in
late-layer representations that more strongly encode task-relevant information \cite{jin2024conceptdepth,tenney2019bert}.
The same ordering is well documented in general language models, where early layers carry surface and syntactic features and task-specific
information concentrates in the middle and late layers \cite{tenney2019bert,jawahar2019bert,lin2019opensesame}, so the pattern seen here for PLMs is
not specific to protein sequence data.

Comparable work in language models suggests a similar preservation-specialization trade-off: with the quantized pretrained backbone frozen and only
low-rank adapters being trained \cite{dettmers2024qlora},
LoRA has been shown to retain more out-of-domain capability than full fine-tuning \cite{biderman2024lora},
consistent with the stronger preservation of early-layer representations observed here.
LoRA can also introduce new task-specific, high-ranking singular directions that are often absent when full
fine-tuning \cite{biderman2024lora},
providing a plausible mechanism for the stability of low-level presentations while middle and late layers diverge during task-specific adaptation.

The ESM-2 model family, however, shows higher
heterogeneity. While representations tend to diverge from early to later layers, the 650M model shows a distinctly different
divergence with mid-to-late layers showing substantially lower CKA values whereas the 3B model remains comparatively aligned
through the network middle before settling into a moderate similarity in later layers, more in line with the smaller parameters
models. This suggests that QLoRA and full-fine-tuning agreement is not simply monotonic with model sizing, even when the underlying
model architecture is identical. The newer ESMC models both show divergence earlier in the layer depth as compared to the related ESM-2
family and then maintain moderate similarity for the remaining layers.

The remaining models further suggest that the layer location of full and QLoRA fine-tuning divergence depends on the pretrained model itself rather than following a standard pattern. Ankh Base is a notable exception to the more typical divergence at later layers, showing substantial early and middle layer divergence followed by convergence in the final layers. This may indicate that this specific architecture and pretraining regime reorganizes intermediate sequence features differently between QLoRA and full fine-tuning while arriving at similar final representations. Ankh3 Large, however, follows a pattern closer to the ESM-2 and ESMC families, suggesting that the newer Ankh model behaves more similarly to the larger, newer PLM families. ProtBERT shows high similarity throughout the first half of the layers before diverging in later layers, consistent with the idea that adaptation differences between QLoRA and full fine-tuning occur in the last layers.

When comparing models with the same architecture but very different parameter sizes, ESM-2 8M and 3B (Fig.~\ref{fig:cka}b), an interesting pattern emerges. While the smaller 8M model shows characteristically high CKA similarity in earlier layers and moderate divergence in later layers across different tasks, this cross-task homogeneity largely dissipates at the much larger 3B scale: the solubility model remains nearly identical across all layers, whereas GB1 and GFP fine-tuning show larger, more task-specific differences that emerge earlier in the layer depth. One possible explanation for these observations is that smaller models have a more constrained representational space, causing different tasks to induce similar patterns of QLoRA/full fine-tuning divergence. Larger models, by contrast, may contain a richer representational space that allows different tasks to reshape distinct regions of the representation hierarchy, producing clearer task-specific divergence.

To evaluate how much each layer contributes to prediction, we used kNN task probes \cite{rajani2020knn,alain2016probes}.
Layer-by-layer kNN task probes show how a protein language model gradually groups proteins with similar task-relevant properties, and how well those
groupings can be used to recover the correct labels or values.
For each layer, proteins were embedded in their layer-specific representations, and each protein was assigned a proxy prediction based on the average measured values of its nearest neighbors in that layer-specific representation space. The resulting probe score therefore reflects how well a layer organizes sequences according to the downstream task, with higher scores indicating that nearby proteins in representation space tend to have similar task values.

\begin{figure}[htbp]
    \centering
    \includegraphics[width=\textwidth]{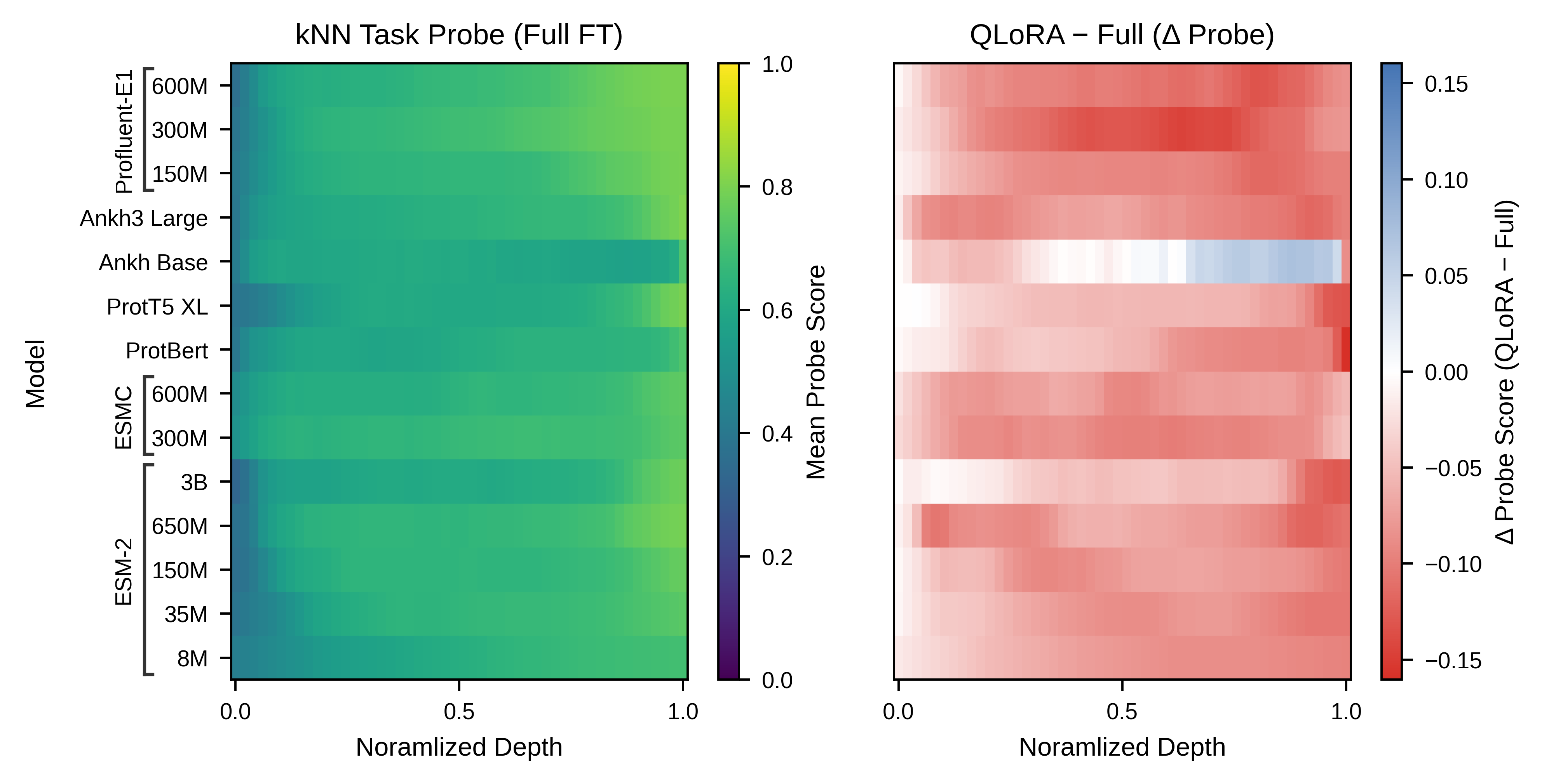}
    \caption{kNN task probe data averaged across six datasets. \textbf{a} Layer-wise task information by kNN probe scores across model depth for full fine-tuned models, averaged across three seeds. \textbf{b} Difference in probe scores between QLoRA and full fine-tuned models, averaged across three seed pairs. Red indicates QLoRA has lower task probe score, blue indicates full fine-tuning has lower task probe score.}\label{fig:probe}
\end{figure}

Across nearly all architectures, both fully fine-tuned (Fig.~\ref{fig:probe}a) and QLoRA fine-tuned models (Fig.~\ref{fig:probe}b) exhibit lower probe scores (blue) in early layers, followed by increasing contributions (green/yellow) from middle and late layers.
This indicates that task-relevant information is weakly encoded in early layers and generally becomes more substantial with model depth, reaching its strongest contributions in later layers, aligning with findings in the broader language modeling umbrella \cite{tenney2019bert,jawahar2019bert,lin2019opensesame}.
This broad depth-dependent pattern is similarly present in the pretrained base models (Fig.~S3a), suggesting that fine-tuning amplifies layer-wise organization of task-relevant information
rather than completely creating it.

There is a notable difference between the ESM2 and ESMC family of models.
The ESM2 family of models all have lower contribution from the initial layers, whereas the ESMC family shows as much contribution from the beginning layers as all layers except the very last (Fig.~\ref{fig:probe}a).
This suggests that the importance of individual layers is dependent both on model architecture. ESMC models expose task relevant structures earlier than other models architectures, with more contribution from early layers than comparably sized models.
Within the ESM-2 and ProfluentE1 families, increasing model size generally shifts task-relevant signal earlier in the layer hierarchy while
preserving strong late-layer contributions, although the largest 3B model shows a marked decrease in the earliest layers (Fig.~\ref{fig:probe}a).
When evaluating the QLoRA PLMs, we notice that while these same patterns are generally preserved in the QLoRA kNN analysis (Fig.~S3a), they exhibit overall
lower probe scores and a relative general muting of the signal as compared to full fine-tuned models (Fig.~\ref{fig:probe}b).
We also observed model-specific patterns of where muted score contributions are most apparent across layer depths. In particular, most models saw the
largest score differences in their later layers when QLoRA fine-tuned, though exceptions like Ankh Base presented higher scores instead.
A lower overall kNN probe score indicates that when fine-tuned using QLoRA, the PLM organizes related proteins less effectively by task-relevant properties,
producing neighborhood-based predictions that are less closely aligned with the ground truth.
Thus, the largest qualitative difference between QLoRA and full fine-tuning is not where task information appears in the network, but rather how strongly that information supports task-specific prediction.

\subsection{Unconditional Generation Quality}\label{subsec:uncond-quality}

\begin{figure}[htbp]
    \centering
    \includegraphics[width=\textwidth]{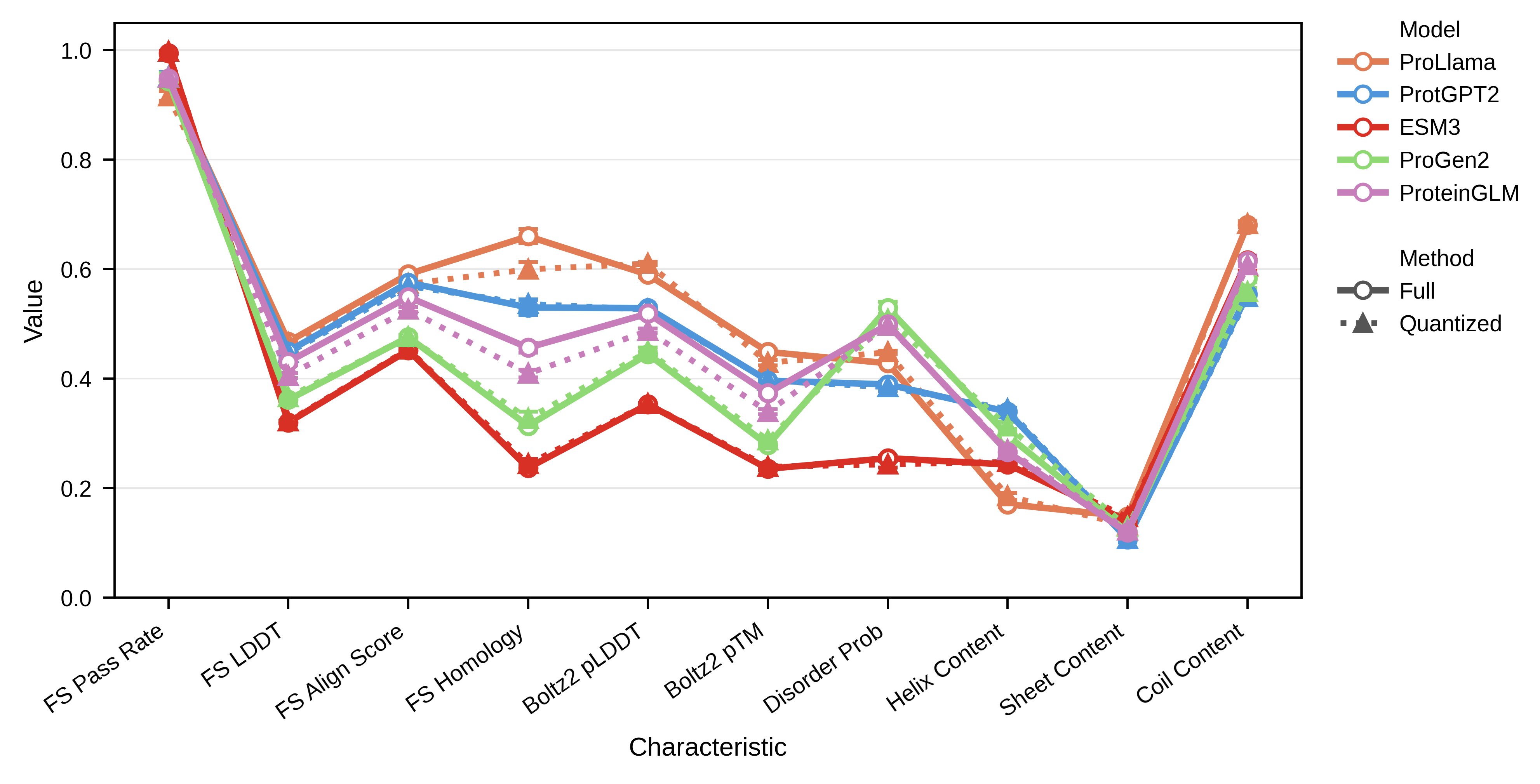}
    \caption{Predicted properties of unconditionally generated sequences from full and quantized models. Profiles compare structural quality, structural similarity, secondary structure content and disorder for each model. Full lines indicate original models, dotted indicate quantized models.}\label{fig:uncond}
\end{figure}

Next, with the goal of understanding how well quantized models can generate novel and realistic proteins, we evaluated five different generative models.
We found all models quantized well with very similar predicted structural properties of the generated proteins (Fig.~\ref{fig:uncond}; Fig.~S4).
Here, each model was tasked with generating 250 sequences per length bin, yielding 750 sequences per model variant per seed. To characterize these hypothetical proteins, the results were evaluated using Foldseek \cite{van2024fast} as described in Section~\ref{subsec:uncond-gen}.
Foldseek metrics for full and quantized variants were nearly identical for all evaluated models, except the homology score was notably decreased for
ProLLaMA and ProteinGLM.

Predicted protein secondary structure content (coil, sheet, helix) of the hypothetical proteins was further evaluated using S4Pred \cite{moffat2021increasing} and predicted disorder using AIUPred \cite{erdHos2024aiupred}.
The results were also practically identical, showing that the average secondary structure content of generated proteins was unaffected by quantization.
Boltz-2 \cite{passaro2025boltz2} predicted pLDDT and pTM of generated sequences, which estimate local structural confidence and global fold accuracy,
were also found to be similar between quantized and full models.
Supporting the potential for PLM quantization in protein generation, we found the full and quantized models were able to generate protein sequences with comparable properties at both the sequence and structure level (Fig.~\ref{fig:uncond}; Fig.~S4).

\subsection{Teacher-Forced KL Divergence}\label{subsec:kl-results}

KL divergence is a useful measure for assessing how quantization changes a model's predicted next-token probability distribution as compared to the
full-precision reference model \cite{kullback1951,gong2024quantization}. This measures how much quantization changes what a protein language model expects to see next by comparing its predicted probabilities for the next amino acid to those of the original model while feeding both the correct sequence.
In this context, a low KL score indicates that the 4-bit model assigns probabilities for next tokens in a manner that is similar to the full model, while high KL indicates quantization substantially alters the model's predictive behavior.
Here, we focus this analysis on the following generative PLMs: ProLLaMA, ProtGPT2, ProGen2, and ProteinGLM.

\begin{figure}[htbp]
    \centering
    \includegraphics[width=\textwidth]{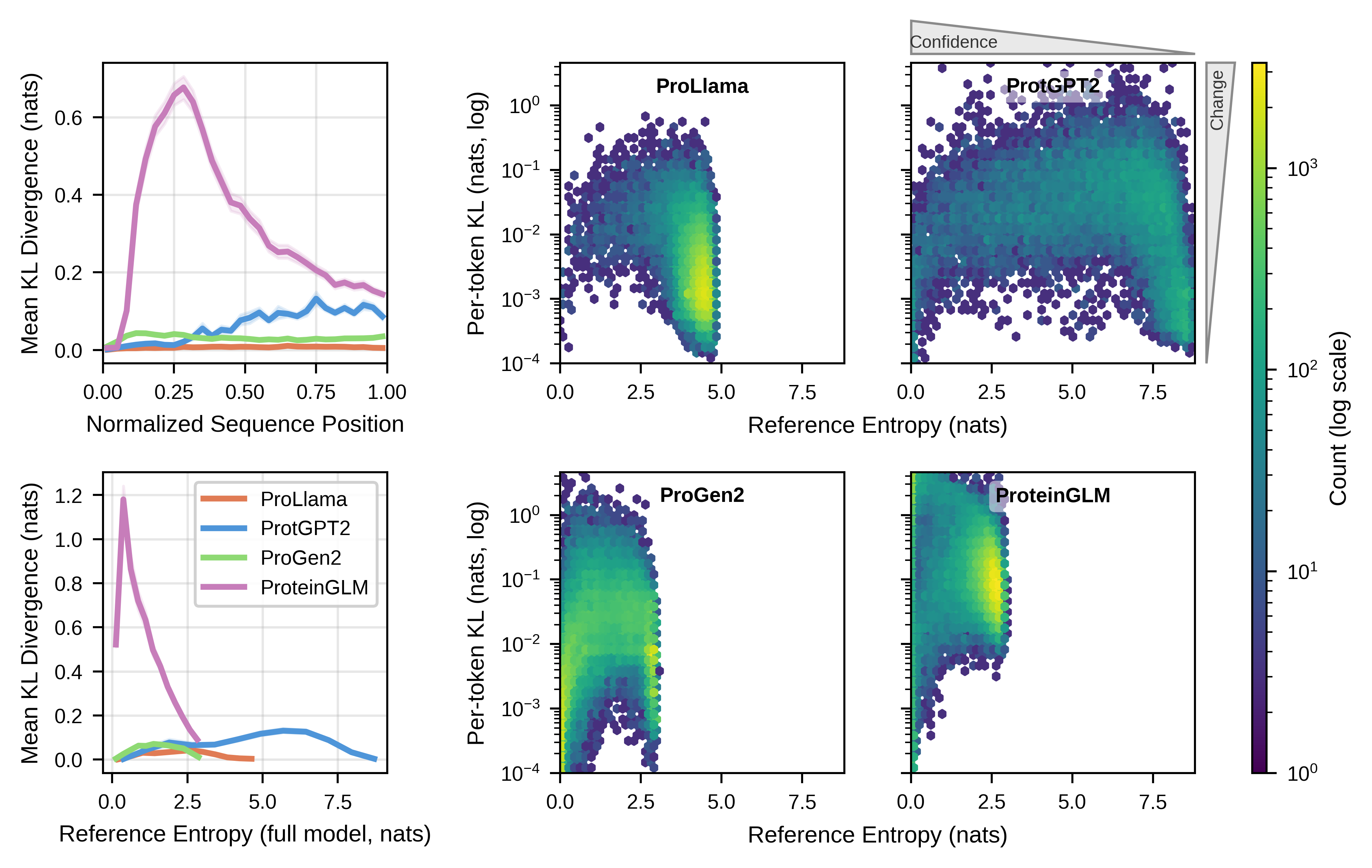}
    \caption{Teacher-forced KL divergence between full and quantized autoregressive models. \textbf{a} Mean token-level divergence as a function of normalized sequence position and \textbf{b} full-model reference entropy. \textbf{c} Joint count density of tokens over per-token KL divergence and full-model reference entropy, showing where quantization-induced discrepancies are concentrated relative to the reference model's confidence levels. Per-token KL is pooled over three generation seeds and three length bins per model. Lines: bin-wise mean $\pm$ 95\% bootstrap CI (1,000 resamples); bins with $<$10 tokens omitted. Hexbins: all tokens, shared log-count scale, axes clipped at the 99.5th percentile.}\label{fig:kl}
\end{figure}

The teacher-forced KL analysis (Fig.~\ref{fig:kl}) shows substantial variation between models in how 4-bit quantization changes next-token distributions. Across normalized sequence position (Fig.~\ref{fig:kl}a), ProLLaMA and ProGen2 show low KL values over the full sequence length, indicating the quantized variants largely preserve the full-precision next-token distributions.
Sequence position was normalized from 0 to 1.
ProtGPT2 (blue) shows a moderate position-dependent increase in KL as sequences increase in length, suggesting that the effects of quantization become more pronounced later in a generation context.
ProteinGLM (pink) is a clear outlier, with markedly higher divergence compared to the other autoregressive models.
Its KL sharply rises after the first few tokens, and decreases steadily towards the end of the sequence.
Overall, ProLLaMA (orange) shows the most consistent next-token distributions under quantization.

We next compared KL divergence against the entropy of the full-precision models' next-token distribution (Fig.~\ref{fig:kl}b). Entropy measures how uncertain a model is about the next token: low entropy means the model strongly prefers a small set of tokens, whereas high entropy indicates probability mass is spread across many tokens. Relating KL to entropy thus helps distinguish whether quantization causes large changes when the reference model is confident or uncertain. This is also useful for interpreting token-level density plots (Fig.~\ref{fig:kl}c): models with low reference entropy and high KL indicate substantial disagreement with the full-precision model, even when the reference-model is confident.

Considering this, ProLLaMA shows the strongest robustness to quantization, with the lowest KL across its entropy range, only a shallow increase at intermediate entropy, and near-zero divergence at higher entropy. ProGen2 similarly has low KL, with only a small peak at low-to-intermediate entropy. ProtGPT2 shows a much broader entropy range than the other models and reaches its highest mean KL at medium-to-high reference entropy. This implies that ProtGPT2 is most affected by quantization when the full-precision model is uncertain, so small numerical effects from quantization may more easily redistribute probability across competing predictions. ProteinGLM again strongly differs from other models, showing highest KL at low reference entropy, followed by a monotonic decrease as entropy increases. This pattern suggests that quantization shifts probability mass away from the full model's preferred tokens in ProteinGLM, especially when the full-precision model is most confident and prefers a narrower set of next tokens.

\subsection{Compute Performance}\label{subsec:compute-results}

\begin{figure}[htbp]
    \centering
    \includegraphics[width=0.93\textwidth]{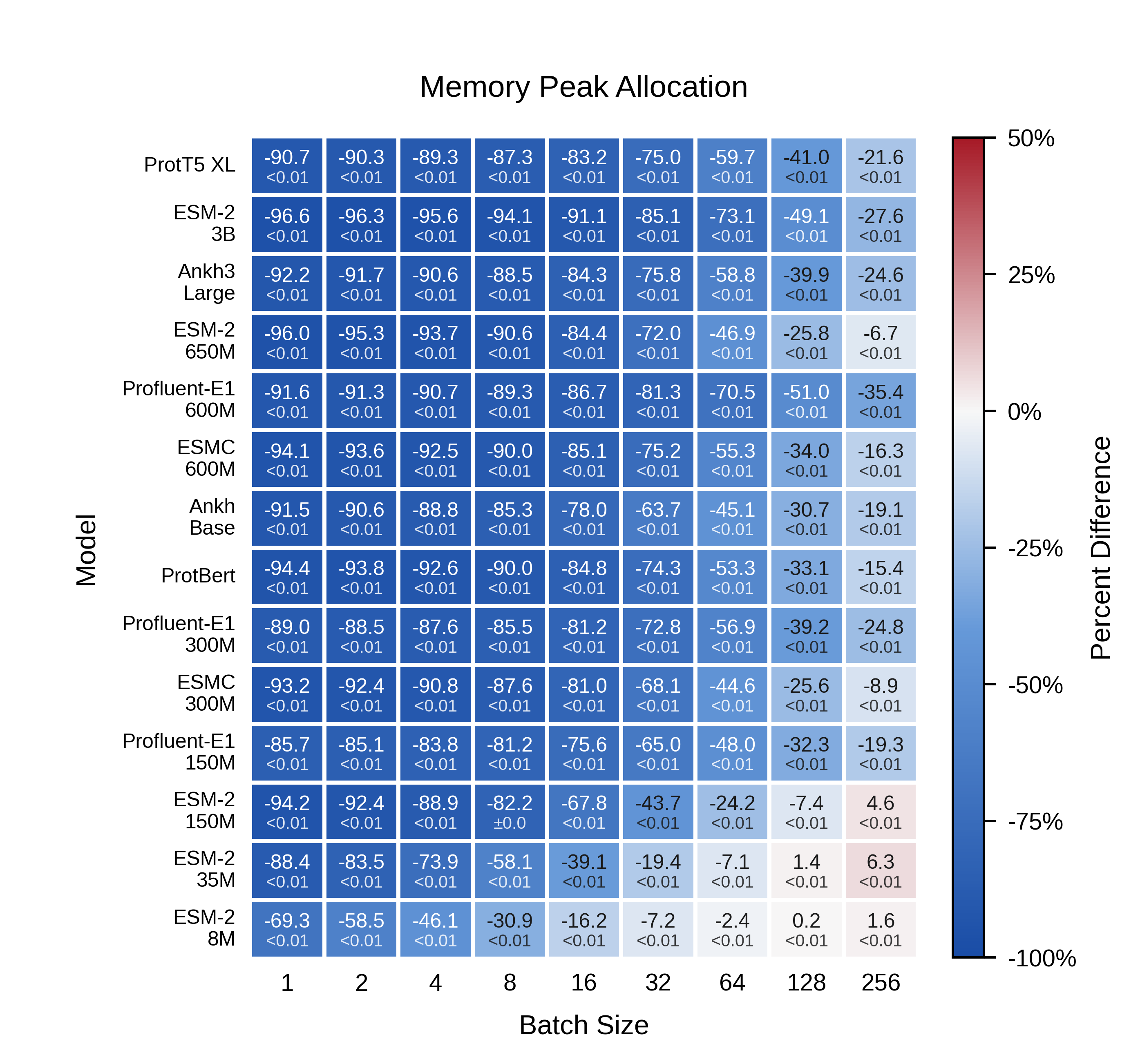}
    \caption{Peak training memory allocation across models and batch sizes. Values show percent differences for QLoRA relative to full-fine-tuning, with negative values indicating reduced memory usage for QLoRA. Standard deviation across three replicates is listed in smaller text.}\label{fig:mem}
\end{figure}

\begin{figure}[htbp]
    \centering
    \includegraphics[width=\textwidth]{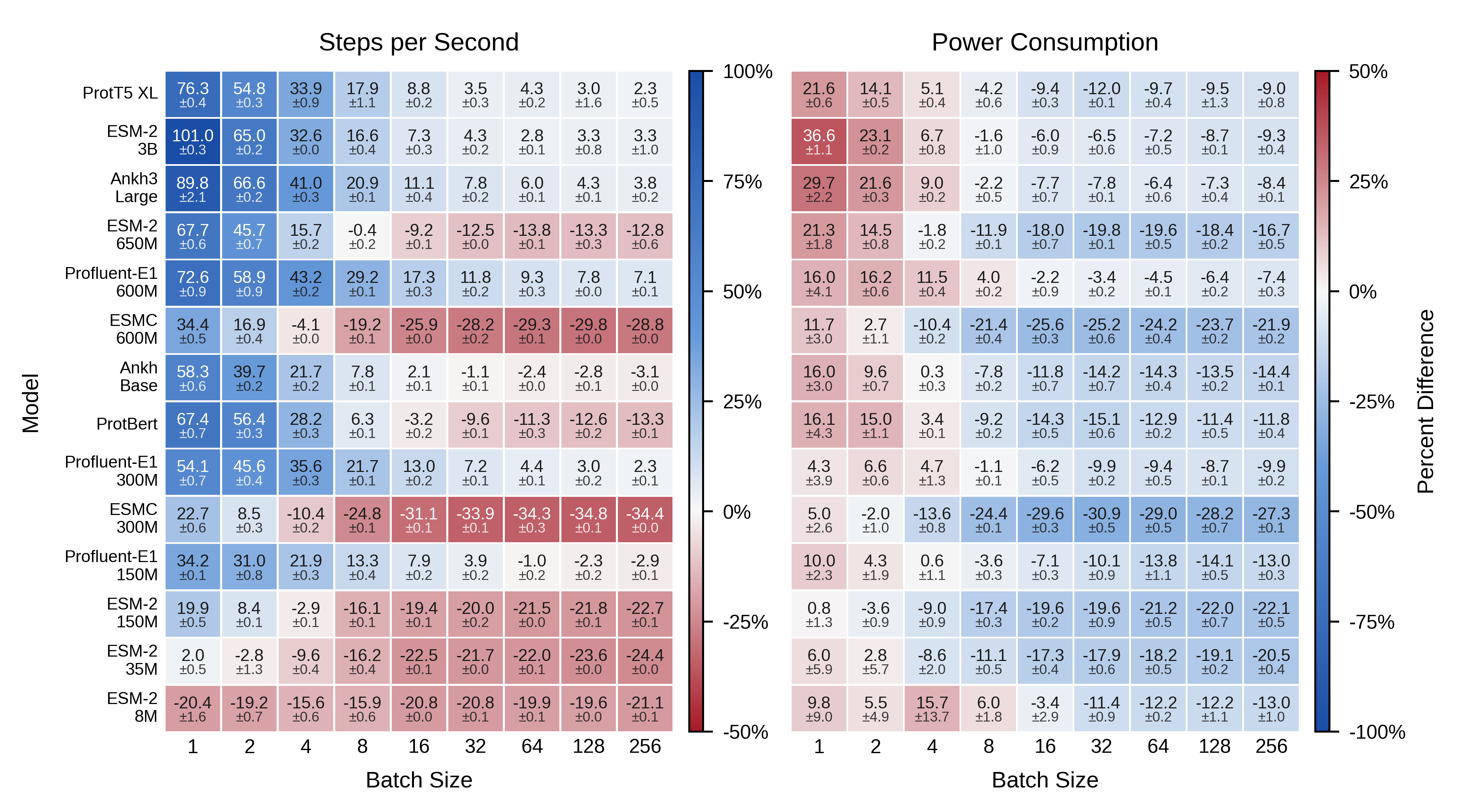}
    \caption{Training throughput and power draw across models and batch sizes. \textbf{a} Percent difference in average training steps per second and \textbf{b} percent difference in average GPU power draw for QLoRA relative to full fine-tuning. All values include standard deviation across three replicates.}\label{fig:throughput}
\end{figure}

\begin{figure}[htbp]
    \centering
    \includegraphics[width=0.5\textwidth]{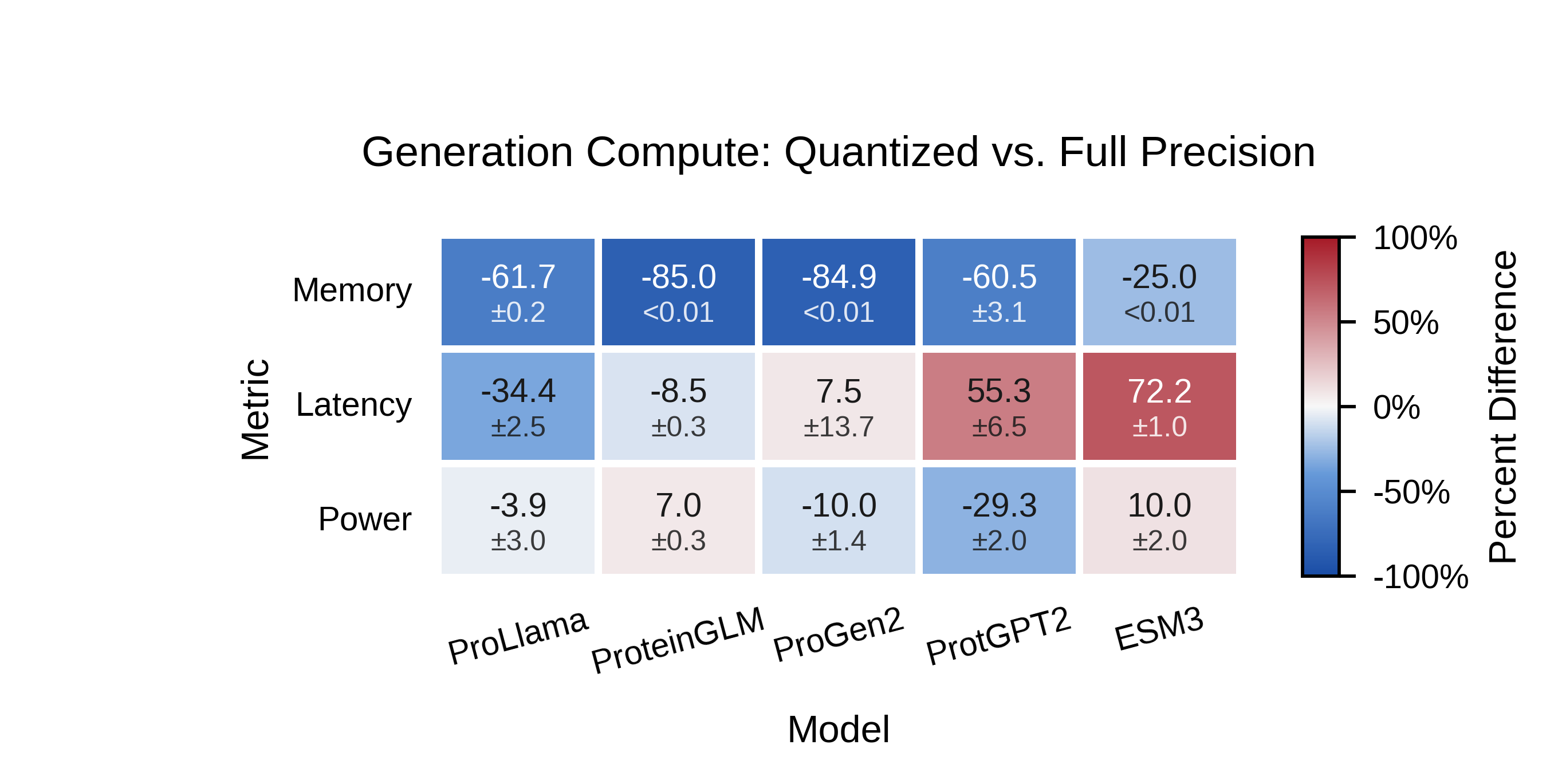}
    \caption{Peak memory, inference latency and average power consumption for unconditional generation.}\label{fig:gen-compute}
\end{figure}

To assess the practical efficiency of 4-bit quantization of PLMs, we measured whether these methods reduced the computational cost of using PLMs during
both fine-tuning and sequence generation.
Specifically, we considered three major metrics of compute cost: evaluated GPU memory usage, average power draw, and training speed (Figs.~\ref{fig:mem}--\ref{fig:gen-compute}).
These metrics determine whether larger models or larger batch sizes can be used within practical hardware constraints \cite{gholami2021survey,shi2025quantchar}.
We evaluated these metrics across 14 different PLMs, ranging from 8 million to up to 3 billion parameters. To further evaluate how best to implement QLoRA, we also tested the influence of batch size on these parameters during fine-tuning by evaluating across 9 batch sizes.

Nearly all models and batch sizes used less memory when quantized, averaging a total of 63\% less memory (Fig.~\ref{fig:mem}).
Larger models benefited the most, with memory usage decreasing by up to 90\% during fine-tuning. Across batch sizes, the greatest
memory reductions were observed at the smallest batch sizes. Some models like Profluent-E1 and ESM-2 3B maintained the memory improvements even at large batch sizes,
whereas other models used more memory than the baseline.

These findings are consistent with previous studies of QLoRA on large language models, which demonstrated that 4-bit quantization substantially reduces GPU memory requirements and enables fine-tuning of models that would otherwise exceed available GPU memory \cite{dettmers2024qlora}.
Our observations are also consistent with application-specific studies reporting that QLoRA substantially reduces GPU memory requirements while maintaining comparable downstream performance, although the exact memory savings depend on the model architecture and software implementation \cite{shakya2025clinical,LoraNatureComm}.
Similar to these reports, we find that the largest memory reductions occur for larger models, while the benefits become more variable across different training configurations.

We also evaluated average power draw during training and found that QLoRA decreased power usage 6\% on average (Fig.~\ref{fig:throughput}b). Unlike memory usage, however, the effect of batch size was flipped: smaller batch sizes consumed more power with QLoRA, while larger batch sizes more often reduced power usage, with a batch size of at least 16 sufficient to improve power consumption while still improving memory consumption for all but the smallest models. Training speed, measured as complete steps per second, increased by 7\% on average with QLoRA (Fig.~\ref{fig:throughput}a). This training speed improvement was less consistent across models, with the ESM-2 and ESMC showing reduced training speed as batch size increased.

Having characterized the compute cost of fine-tuning, we next asked whether the same benefits hold when quantization is used on its own, without adapters.
We therefore evaluated these properties (memory, training speed, power) for each of the five generative models (Fig.~\ref{fig:gen-compute}), which are
quantized for inference rather than fine-tuned and so form a separate comparison from Figs.~\ref{fig:mem} and \ref{fig:throughput}.
This setting tests the cost of using quantization alone for de novo sequence generation.
Quantization substantially reduced peak memory GPU usage during unconditional generation relative to full-precision models.
This remains consistent with literature findings, where reducing weight precision lowers the memory footprint of inference roughly in proportion to the bit width, making models runnable on hardware that could not otherwise host them \cite{gholami2021survey,shi2025quantchar}.

The effect of quantization however was more varied in its impact on latency and power (Fig.~\ref{fig:gen-compute}).
All generative models showed reductions in memory usage, with the largest reductions for ProGen2 and ProteinGLM.
ProLLaMA and ProteinGLM showed reduced latency, whereas the remaining models,
especially ESM3 and ProtGPT2, exhibited increased latency with quantization.
This observation aligns with prior work in LLMs, where memory bottlenecks can be addressed with quantization at the cost of model performance \cite{gong2024quantization,jacob2018quantization}.
Power usage with quantization was decreased for ProLLaMA, ProtGPT2 and ProGen2, but
increased for ESM3 and ProteinGLM. ProLLaMA was the only model that exhibited
improvements across all three metrics: memory, latency and power. Across all metrics
and models, variance was generally low, except for ProGen2 latency.
It has been demonstrated that differences in model size, architecture, and task cause both increased and decreased latency and energy efficiency \cite{shi2025quantchar}.
While we only evaluated generative models for a single task (unconditional protein generation), our results show that some models are better suited regarding latency or power usage.
For any use case involving generative PLMs, model selection should therefore consider the specific task and use case, and be thoroughly evaluated, perhaps more so than in simple fine-tuning.

\section{Discussion}\label{sec:discussion}

We establish a practical framework for the quantization and adaptation of protein language models (PLMs) across diverse protein modeling use cases.
Across different protein prediction tasks, converting the majority of model weights to 4-bit precision and fine-tuning with
LoRA substantially reduced memory requirements while many model-task pairs retained approximately 90-97\% of full fine-tuning
performance (Fig.~\ref{fig:ft-ratios}). In some cases, QLoRA fine-tuned models matched or exceeded the performance of their full-precision counterparts.
These results suggest that, for many common protein property prediction tasks, task-specific adaptation can be captured
by low-rank updates on a quantized pretrained model rather than requiring updates to all model parameters. The practical
benefit was largest in computational cost: peak memory usage was reduced by an average of 63\%, with reductions approaching
90\% for larger models (Fig.~\ref{fig:mem}). This makes QLoRA particularly relevant for bioinformatics groups with limited access to high-memory
accelerators. The specific computational benefits of QLoRA were, however, not uniform across PLM architectures, datasets, or training settings.

ESMC models showed strong downstream performance and, in some cases, performed on par with substantially larger ESM-2 models, consistent with findings
from Cándido et al., where representations generated from ESMC models enabled highly accurate prediction of protein interaction tasks \cite{candido2026worldmodel}.
In contrast, T5-based models were much more sensitive to hyperparameters and were generally more difficult to fine-tune effectively with QLoRA.
While there is some evidence that Ankh models fine-tune less effectively than ESM models in general, and not only under quantization \cite{LoraNatureComm},
it is more likely that this is related to the T5 model's span-corruption pre-training objective \cite{raffel2020t5}, which differs from the masked-language-modelling objective used by the encoder-only models evaluated here (ESM-2, ESMC, ProtBERT and Profluent-E1) \cite{devlin2019bert}.
Task-level behavior also varied: protease stability
prediction was consistently more challenging than fluorescence brightness prediction or token-level secondary-structure
prediction.

Computational benefits were similarly dependent on model and batch size. Although all models showed reduced
memory utilization at small batch sizes, this advantage decreased for larger batch sizes and, in some cases, QLoRA models
required more memory than their full-precision counterparts. This is consistent with previous reports \cite{dettmers2024qlora},
where larger batch sizes increase the memory footprint for standard LLMs. Together, these results show that QLoRA should
not be viewed as an architecture-agnostic replacement for full fine-tuning. Model family, task type, hyperparameter tuning,
LoRA configuration, and the choice of model layers to quantize remain important determinants for successful implementation of QLoRA fine-tuning.

These findings support a practical workflow for supervised protein prediction tasks under constrained computational resources
(Table~\ref{tab:workflow}, Table~S2). With this in mind, we recommend treating QLoRA as a first-pass strategy when GPU memory limits make full
fine-tuning difficult, especially for larger PLMs where memory savings are the largest. Rather than replacing full fine-tuning outright,
QLoRA can be used as an initial screen, with full fine-tuning reserved for cases where QLoRA does not recover sufficient validation
performance or where the target task is known to be difficult. When feasible, practitioners should compare QLoRA with full
fine-tuning on a representative validation subset, report the QLoRA-to-full performance ratio with uncertainty, and proceed
with QLoRA when performance recovery is high relative to computational savings. More extensive validation is warranted for
tasks with weak baseline performance, high variance or known sensitivity to learning rate and batch size.

Representation analyses (Fig.~\ref{fig:probe},\ref{fig:kl}) help explain the frequent parity between QLoRA and full fine-tuning performance while still producing
model-specific differences. CKA showed that early-layer representations were generally preserved between QLoRA and full
fine-tuning, consistent with the idea that low-level sequence representations remain mostly intact. Divergence between
fine-tuning regimes was typically concentrated in middle and late layers, where task-specific adaptation is expected to
occur \cite{nadipalli2025layerwise,jin2024conceptdepth}.
This mirrors the layer-wise picture established in general language models, where probing studies place surface features in early layers and
task- and semantics-specific information progressively deeper in the network \cite{tenney2019bert,jawahar2019bert,lin2019opensesame},
and where fine-tuning reshapes late-layer representations while leaving early layers largely intact \cite{nadipalli2025layerwise}.

kNN task probes further showed that both full and QLoRA fine-tuning encode task-relevant structure increasingly in later layers,
although probe scores were generally weaker with QLoRA. These results support the view that QLoRA
preserves much of the broad representational geometry of PLMs, but often induces a lower magnitude task adaptation
than full fine-tuning. Similar findings have been reported for general language models, where low-rank adaptation learns less of the target task than
full fine-tuning but correspondingly forgets less of the pretrained distribution, a trade-off attributed to the restricted rank of the update \cite{biderman2024lora}.

For unconditional protein generation, quantization had model-dependent effects that were not always evident by aggregate sequence-level metrics. ProtGPT2,
ESM3 and ProGen2 generally retained predicted structural and sequence level properties similar to their full-precision counterparts, whereas ProteinGLM
and ProLLaMA showed minor deviations in their radar plot summaries, particularly in homology probabilities to natural sequences (Fig.~\ref{fig:uncond}).
However, teacher-forced KL analysis of the autoregressive models revealed that similar aggregate properties can mask substantial differences in next-token
distributions. ProLLaMA and ProGen2 maintained low divergence between quantized and full-precision next-token distributions, while ProtGPT2 showed moderate position- and entropy-dependent divergence. ProteinGLM was a clear outlier, with high divergence especially at low reference entropy, indicating that quantization substantially changed its next-token behavior when the reference full-precision model was most confident.

These generative results suggest that quantization might be useful for reducing memory requirements during protein sequence generation,
but should be validated with distribution-level diagnostics. ProteinGLM, for example, still produced sequences with predicted properties
broadly similar to the full model while showing lower homology probabilities and much higher KL divergence. This could indicate that
quantization introduces divergence from known proteins, but it also implies that the quantized model should not be assumed to sample
from the same learned distribution as the full-precision model. Interestingly, reduced homology probability was also observed for
quantized ProLLaMA, yet it exhibited the lowest and most consistent KL divergence. This suggests that even a small token probability
shift (indicated by low KL divergence) can change the overall profile of the generated proteins. From a compute efficiency perspective,
all generative models used substantially less memory under quantization, but only ProLLaMA showed improvement across memory, latency
and power usage. For quantized generative PLMs, we therefore recommend pairing standard predicted structural and sequence-level
metrics with distributional diagnostics such as teacher-forced KL divergence, entropy-stratified divergence, token frequencies
and length distributions.

This study has several limitations. Generated proteins were assessed using computational predictors and structural
search rather than experimental validation. KL divergence analysis was restricted to autoregressive models and was
not applicable to models like ESM3 that use iterative masked decoding. Additionally, the architecture-specific behavior
observed here means these conclusions may not transfer directly to future PLMs or to different quantization and
parameter-efficient fine-tuning methods. Hardware differences may also affect performance, particularly because support
for INT4 computation varies across GPU platforms, including Ada, Hopper and Blackwell architectures. Prior research has shown that
different GPU architectures can affect quantization significantly,
exhibiting higher latency or scaling effects depending on the model \cite{shi2025quantchar}.

Overall, our findings support QLoRA as a practical approach for reducing the computational cost of adapting large PLMs,
and support quantization as a useful tool for lowering the memory cost of generative PLMs. At the same time, these benefits
are strongly model- and task-dependent. In practice, we recommend treating QLoRA as the first strategy to evaluate when
limited by GPU memory, followed by targeted validation against full fine-tuning when feasible. For generative use cases,
quantized models should be evaluated not only with aggregate predicted properties but also with distributional diagnostics
capable of detecting shifts in next-token behavior. Despite the limitations mentioned above, this study provides practical
evidence that quantization and low-rank adaptation can substantially reduce computational requirements while broadly preserving
the capabilities of PLMs for supervised task adaptation and unconditional sequence generation.
We see significant advantage in settings like automated discovery workflows, where rapid prediction can be
augmented with QLoRA.


\begin{table}[htbp]
    \caption{Suggested workflows for quantized PLM adaptation}\label{tab:workflow}
    \centering
    \begin{tabular}{@{}>{\RaggedRight\arraybackslash}p{2.8cm}>{\RaggedRight\arraybackslash}p{2.8cm}>{\RaggedRight\arraybackslash}p{3.0cm}>{\RaggedRight\arraybackslash}p{3.0cm}@{}}
        \toprule
        \textbf{Use case}                      & \textbf{Recommended starting point}                                  & \textbf{Validation check}                                                                          & \textbf{Signals for caution}                                                                                                    \\
        \midrule
        Supervised protein property prediction & Start with QLoRA (especially when GPU memory is limiting)            & Compare QLoRA against full fine-tuning on representative validation subset (ratio and uncertainty) & QLoRA recovery below practical threshold, typically $<$90-95\%*, large uncertainty introduced or latency degrades               \\
        \addlinespace
        Low signal or high variance tasks      & Use QLoRA as initial screen, not as a default                        & Run multiple seeds and compare absolute performance and variance, not just QLoRA/full ratio        & Baseline full fine-tuning performance is weak/highly variable                                                                   \\
        \addlinespace
        Hyperparameter sensitive models        & Perform a small learning rate and batch size sweep before comparison & Check convergence stability across seeds                                                           & Results are unstable across random seeds, learning rates or batch sizes                                                         \\
        \addlinespace
        Unconditional generation               & Use 4-bit quantization for exploratory memory efficient generation   & Compare sequence-level and structural metrics between full and quantized models                    & Token-level or entropy stratified KL shifts substantially; consider residue frequencies, length distribution, diversity metrics \\
        \addlinespace
        Downstream experimental design         & Consider quantization as a bonus, not as the only decision factor    & Confirm quantized outputs preserve relevant design objectives                                      & Generated candidates will be experimentally validated                                                                           \\
        \bottomrule
    \end{tabular}
    \par\vspace{4pt}
    {\small\raggedright * The 90-95\% recovery threshold is intended as a practical heuristic rather than a universal threshold; stricter thresholds may be appropriate for high-cost experimental or safety critical applications.\par}
\end{table}

\section{Conclusions}\label{sec:conclusions}

We have shown that PLMs can often be quantized and efficiently fine-tuned to substantially reduce
computational cost without large losses in downstream performance. Across diverse prediction tasks,
QLoRA greatly reduced memory footprints while generally preserving performance, often retaining more
than 90\% of full fine-tuning performance and the largest models showing the greatest memory savings.
These results support QLoRA as a practical strategy for adapting large PLMs when GPU memory is a limiting
factor, while also showing that performance depends on model family, dataset and training configuration. QLoRA 
may therefore broaden access to PLMs for protein structure and discovery applications, particularly in settings with limited computational resources

Representation analysis further showed that QLoRA often preserves early-layer structure while inducing task-specific changes
in later layers, broadly mirroring full fine-tuning albeit with reduced magnitude of representational change. For unconditional
generation, 4-bit quantization produced broadly similar predicted sequence and structural properties for multiple generative
PLMs compared with their full-precision counterparts. However, teacher-forced KL divergence revealed substantial model-dependent
shifts in autoregressive output distributions, indicating that predicted sequence quality alone may not fully capture the effects
of quantization on generative models.

Overall, our results support QLoRA and 4-bit quantization as practical tools for reducing computational requirements of PLM
fine-tuning and deployment, particularly GPU memory usage. These findings support a practitioner workflow in which QLoRA is used
as a first-pass strategy for supervised PLM adaptation under memory constraints, followed by targeted comparison with full fine-tuning
when feasible, especially for challenging tasks, unstable architectures or low validation recovery. For quantized generative PLMs,
standard sequence-level and structural evaluations should be paired with distributional diagnostics before using generated sequences
in downstream applications. By lowering resource barriers, these approaches can make large-scale PLMs more feasible to use on commonly
available and lower-cost GPUs, while still requiring model- and task-specific validation when predictive performance or generative distributions are critical.

\section*{List of Abbreviations}
\noindent
ML: Machine learning\\
PLM: Protein language model\\
LLM: Large language model\\
PEFT: Parameter-efficient fine-tuning\\
LoRA: Low-rank adapters\\
QLoRA: Quantization and low-rank adapters\\
GFP: Green fluorescent protein\\
GPU: Graphics processing unit\\
INT4: 4-bit integer precision\\
LR: Learning rate\\
BS: Batch size\\
CKA: Centered kernel alignment\\
kNN: k-nearest neighbor\\
SpearmanR: Spearman's rank correlation coefficient\\
PDB: Protein Data Bank\\
pLDDT: Predicted local distance difference test\\
TM: Template modelling\\
pTM: Predicted template modelling\\
KL: Kullback-Leibler

\section*{Declarations}

\bmhead{Data availability}
The data underlying the results reported in this study are
available from the corresponding author upon reasonable request.

\bmhead{Code availability}
The custom training and evaluation code used in this study is
available from the corresponding author upon reasonable request.

\bmhead{Acknowledgements}
We also thank Dr.\ Krishnan, Dr.\ Cuperlovic-Culf, Dr.\ Leibovich and Sahil Patel for their additional
feedback and for helpful conversations on this work.

\bmhead{Funding}
This work was supported by funds to K.P. from the National Research Council AI4Design program
(NRC File 2024-044); the Canada Research Chairs Program (Files 950-231075 and 950-233107); and the
NSERC Discovery Grants Program (RGPIN-2016-06352).

\bmhead{Competing interests}
The authors declare that they have no competing interests.

\bmhead{Author contributions}
S.C. and I.Y.Z. contributed equally to this work in writing and analysis and are designated co-first authors. P.B. contributed to the initial idea and project direction. 
S.R., I.K. M.X., V.W., S.P. and S.B. contributed to dataset curation, preliminary analysis, and model selection. S.S. provided advice and discussion about results as well as feedback on the draft. K.P. is the supervising author.

\bmhead{Supplementary information}

Supplementary Figures~S1--S5 and Supplementary Tables~S1--S2 are provided in the accompanying
Supplementary Information file.

\bibliography{references}

\end{document}


\title[Supplementary Information]{Supplementary Information for Analysis of Quantized and Efficiently Adapted Protein Language Models}

\author[1]{\fnm{Ilan Yaniv} \sur{Zeisler}}
\equalcont{These authors contributed equally to this work.}

\author[1]{\fnm{Sebastian} \sur{Clancy}}
\equalcont{These authors contributed equally to this work.}

\author[1]{\fnm{Pouriya} \sur{Bayat}}

\author[4]{\fnm{Saaim} \sur{Raad}}

\author[3]{\fnm{Ivan} \sur{Kraskov}}

\author[3]{\fnm{Matthew} \sur{Xie}}

\author[1]{\fnm{Vivian} \sur{White}}

\author[2]{\fnm{Spencer} \sur{Perkins}}

\author[1]{\fnm{Serena} \sur{Singh}}

\author[5]{\fnm{Sepehr} \sur{Bayat}}

\author*[1,2]{\fnm{Keith} \sur{Pardee}}\email{keith.pardee@utoronto.ca}

\affil[1]{\orgdiv{Leslie Dan Faculty of Pharmacy}, \orgname{University of Toronto}, \orgaddress{\city{Toronto}, \state{ON}, \country{Canada}}}
\affil[2]{\orgdiv{Department of Mechanical and Industrial Engineering}, \orgname{University of Toronto}, \orgaddress{\city{Toronto}, \state{Ontario}, \country{Canada}}}
\affil[3]{\orgdiv{Department of Computer Science}, \orgname{University of Toronto}, \orgaddress{\city{Toronto}, \state{Ontario}, \country{Canada}}}
\affil[4]{\orgdiv{Department of Electrical and Computer Engineering}, \orgname{University of Toronto}, \orgaddress{\city{Toronto}, \state{Ontario}, \country{Canada}}}
\affil[5]{\orgdiv{Department of Computing and Software}, \orgname{McMaster University}, \orgaddress{\city{Hamilton}, \state{ON}, \country{Canada}}}

\maketitle

\newpage
\begin{longtable}{@{}>{\RaggedRight\arraybackslash}p{2.6cm}>{\RaggedRight\arraybackslash}p{4.2cm}>{\RaggedRight\arraybackslash}p{5.2cm}@{}}
    \caption{Practical validation and reporting checklist for quantized PLM workflows.}\label{tab:checklist}                                   \\
    \toprule
    \textbf{Workflow}        & \textbf{Checklist item}                                        & \textbf{Purpose/what to report}                                                                                          \\
    \midrule
    \endfirsthead
    \multicolumn{3}{@{}l}{\itshape Table \thetable\ (continued)}                                                                                                                                                         \\
    \toprule
    \textbf{Workflow}        & \textbf{Checklist item}                                        & \textbf{Purpose/what to report}                                                                                          \\
    \midrule
    \endhead
    \midrule
    \multicolumn{3}{r@{}}{\itshape continued on next page}                                                                                                                                                               \\
    \endfoot
    \bottomrule
    \endlastfoot
    Supervised fine-tuning   & Report pretrained model checkpoint and parameter scale         & Enables comparison across models, parameter sizes                                                                        \\
    \addlinespace
                             & Report quantization configuration                              & Bit width, quantization method, target modules, compute dtype                                                            \\
    \addlinespace
                             & Report LoRA configuration                                      & Rank, alpha, dropout, target modules                                                                                     \\
    \addlinespace
                             & Report fine-tuning hyperparameters                             & Learning rate, optimizer, batch size, epoch count/early stopping criterion, random seeds                                 \\
    \addlinespace
                             & Compare QLoRA with full fine-tuning when feasible              & Use same train/validation/test split                                                                                     \\
    \addlinespace
                             & Report absolute performances                                   & Avoid relying only on QLoRA/full performance ratio                                                                       \\
    \addlinespace
                             & Report QLoRA/full performance recovery                         & Include uncertainty across seeds                                                                                         \\
    \addlinespace
                             & Report computational metrics                                   & Peak GPU memory (and latency, power usage if relevant)                                                                   \\
    \addlinespace
                             & Flag low recovery cases                                        & Note cases below predefined practical threshold, e.g.\ 90--95\% recovery                                                 \\
    \addlinespace
                             & Flag low-performing or high-variance tasks                     & Check absolute performance instead of relying only on relative recovery                                                  \\
    \midrule
    Unconditional generation & Report generation setup                                        & Include model checkpoint, quantization settings, sampling method                                                         \\
    \addlinespace
                             & Compare using structural and sequence-level metrics            & Include metrics such as structural confidence, similarity, any application relevant metrics                              \\
    \addlinespace
                             & Compare sequence distributions                                 & Include length and token frequency distributions                                                                         \\
    \addlinespace
                             & Compute token-level distributional diagnostics (when feasible) & For autoregressive models, report teacher-forced KL divergence between full and quantized next-token distributions       \\
    \addlinespace
                             & Stratify diagnostics                                           & Examine by sequence position, sequence length and full model entropy                                                     \\
    \addlinespace
                             & Flag hidden distributional shifts                              & Note cases where aggregate structural metrics significantly change or are preserved but token-level distributions shift  \\
    \addlinespace
                             & Match validation extent to use case                            & Use stricter validation when generated or predicted candidates are used in costly experimental work or pose safety risks \\
\end{longtable}

\begin{figure}[htbp]
    \centering
    \includegraphics[width=\textwidth,height=0.85\textheight,keepaspectratio]{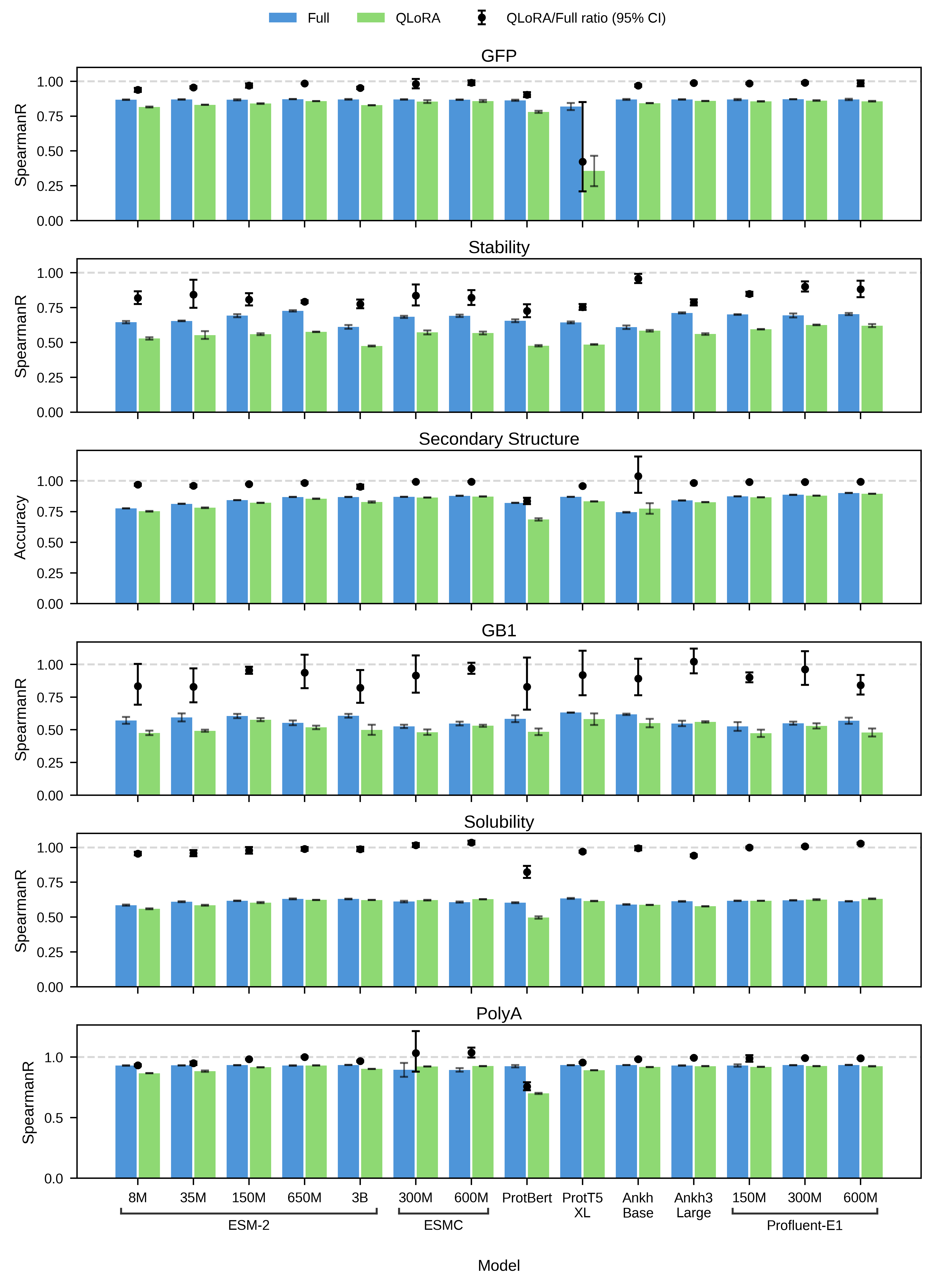}
    \caption{Full fine-tuning and QLoRA performance across downstream tasks. Bars show absolute task performance, and points show QLoRA/full fine-tuning ratios with 95\% confidence intervals.}\label{fig:s1}
\end{figure}

\begin{figure}[htbp]
    \centering
    \includegraphics[width=\textwidth]{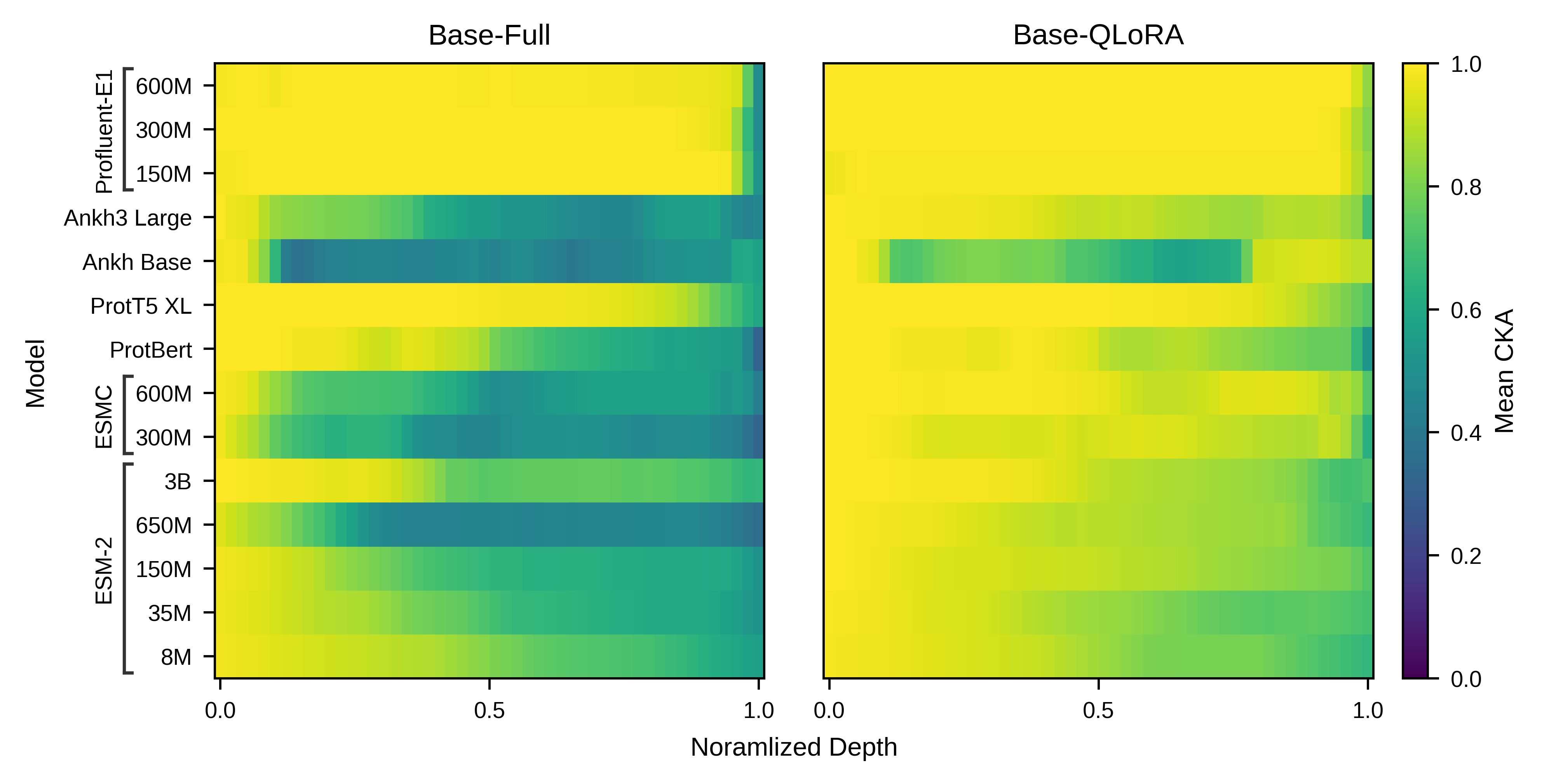}
    \caption{Layer-wise representational similarity to the pretrained base model with \textbf{a} full fine-tuned models and \textbf{b} QLoRA fine-tuned models.}\label{fig:s2}
\end{figure}

\begin{figure}[htbp]
    \centering
    \includegraphics[width=\textwidth,height=0.85\textheight,keepaspectratio]{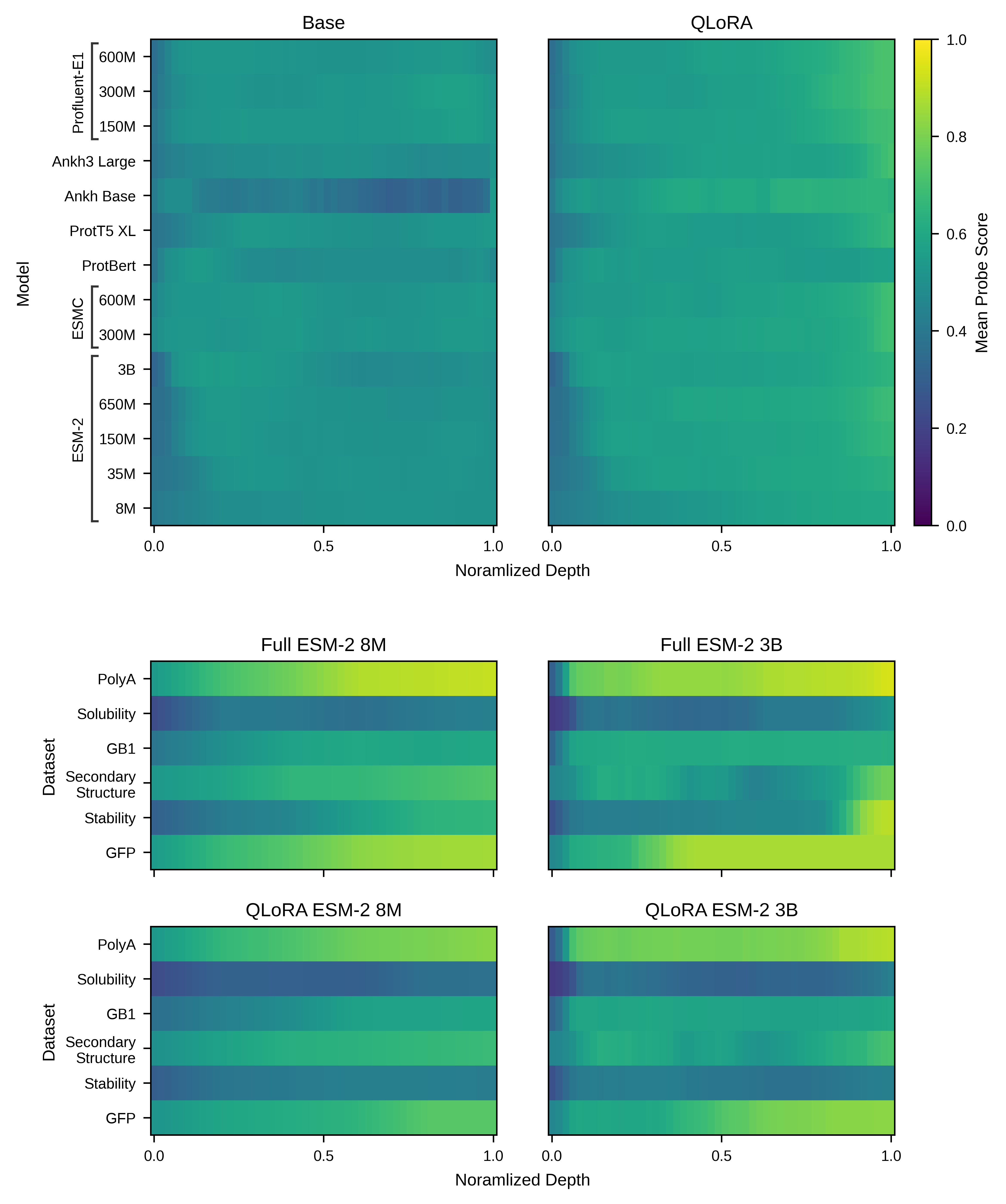}
    \caption{Layer-wise task information in pretrained and ESM-2 fine-tuned representations. \textbf{a} Mean kNN probe scores across normalized depth for pretrained base models and QLoRA fine-tuned models. \textbf{b} Dataset-specific kNN probe scores across normalized depth for ESM-2 8M and 3B after full fine-tuning and QLoRA fine-tuning.}\label{fig:s3}
\end{figure}

\begin{figure}[htbp]
    \centering
    \includegraphics[width=\textwidth,height=0.85\textheight,keepaspectratio]{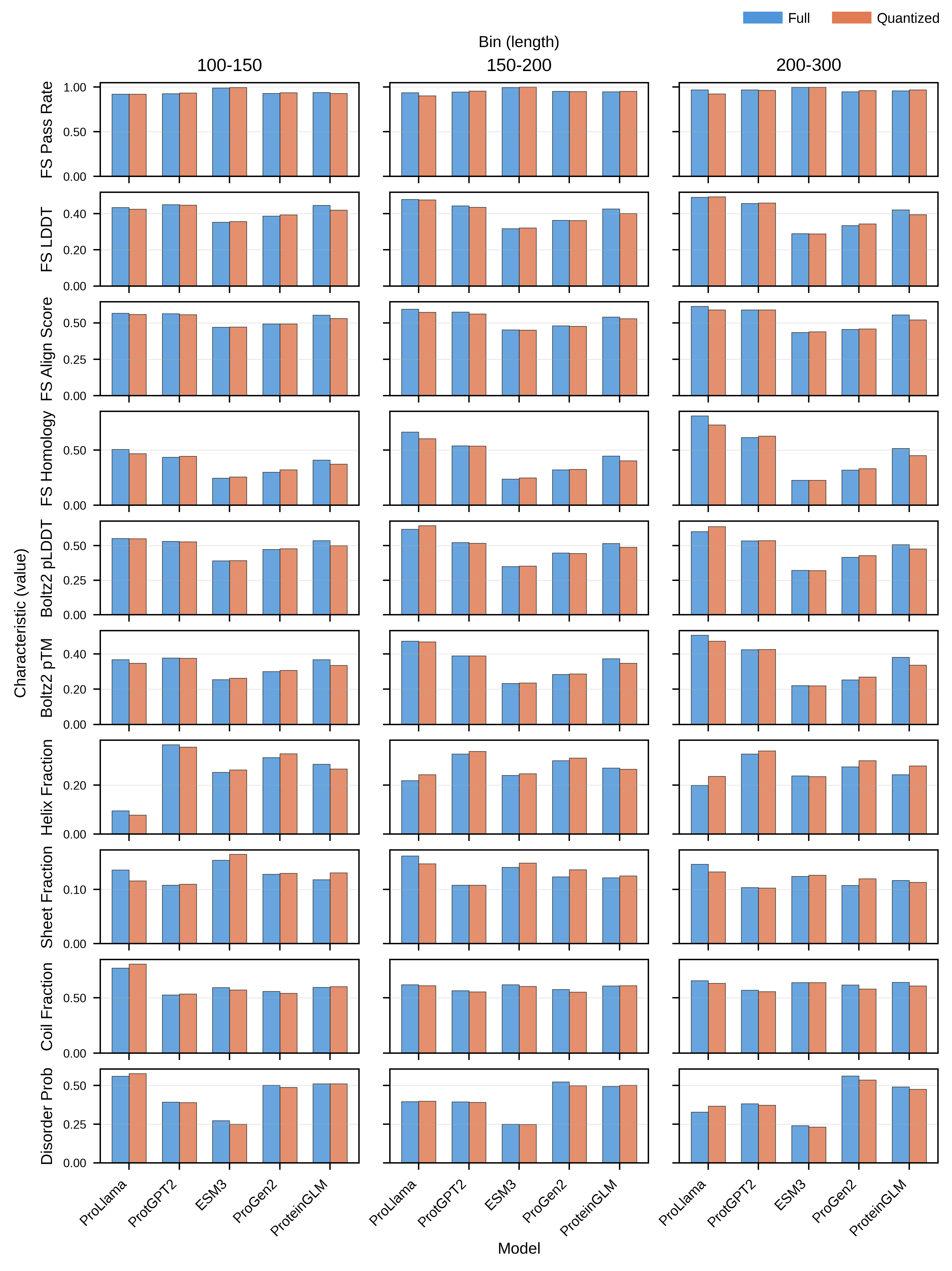}
    \caption{Predicted properties of unconditionally generated sequences from full and quantized models stratified by sequence length bins. Bars compare full and quantized models across Foldseek, Boltz-2, secondary structure and disorder metrics.}\label{fig:s4}
\end{figure}

\begin{figure}[htbp]
    \centering
    \includegraphics[width=\textwidth]{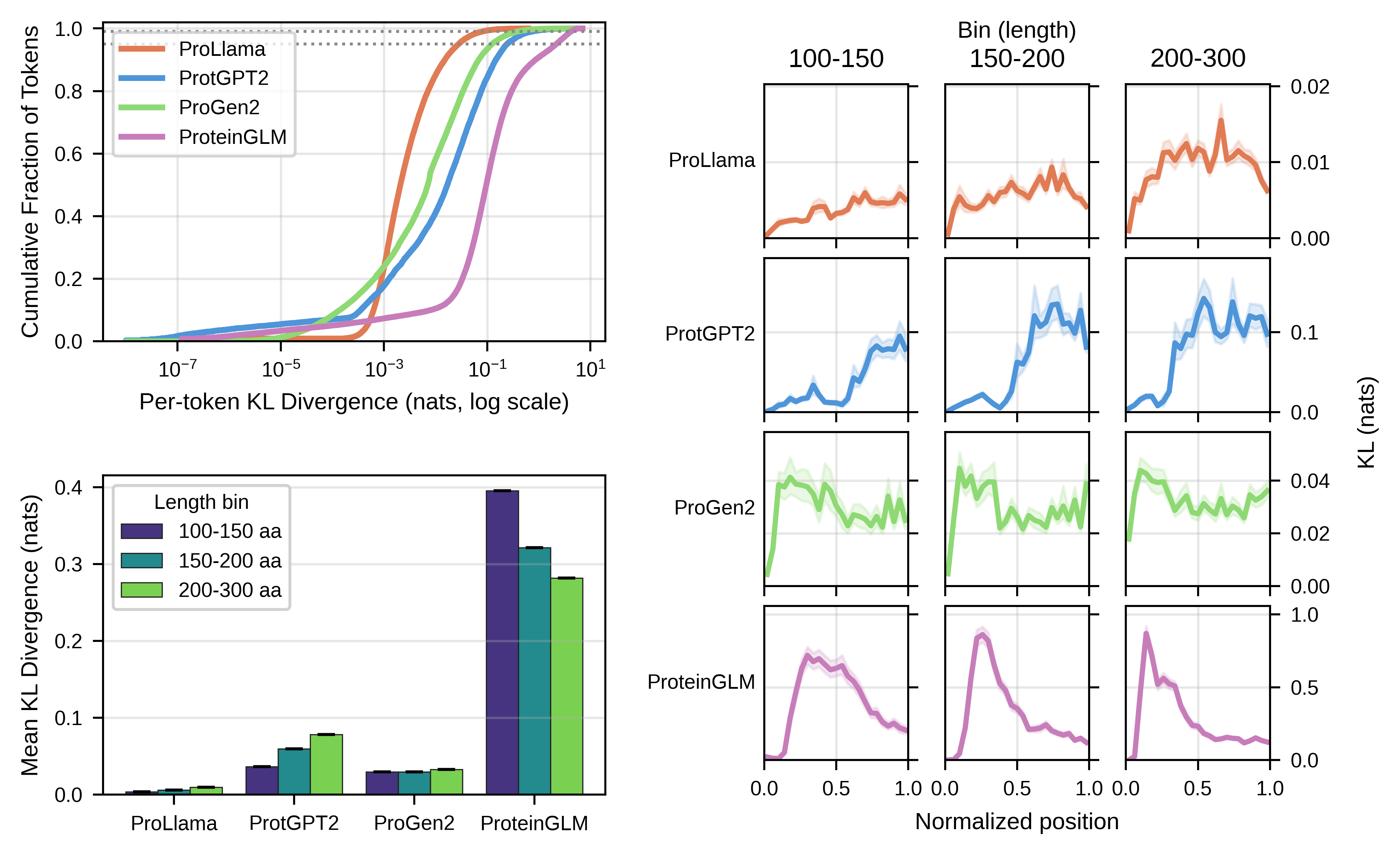}
    \caption{Additional teacher-forced KL divergence analyses. \textbf{a} Cumulative distribution of per-token KL divergence for each autoregressive model. Dashed lines indicate 95\% and 99\% of accumulated tokens. \textbf{b} Mean per-token divergence by model and sequence-length bin. \textbf{c} Mean per-token divergence across normalized sequence position, stratified by model and sequence-length bin.}\label{fig:s5}
\end{figure}


\clearpage
\begin{table}[htbp]
    \caption{Batch size and learning rate for full and QLoRA models fine-tuned on fluorescence, stability and secondary structure datasets. LoRA parameters were fixed with $r=8$, $\alpha=32$, $\mathrm{dropout}=0.05$. Runs were completed on NVIDIA GeForce RTX 4090s, RTX 5090s and a DGX Spark (GB10 Grace Blackwell) depending on model size and availability.}\label{tab:fine-tuning-hyperparams}
    \centering
    \setlength{\tabcolsep}{2pt}\scriptsize
    \begin{tabular}{|c|c|c|c|c|c|c|c|c|c|c|c|c|} \hline
                        & \multicolumn{4}{|c|}{Fluorescence} & \multicolumn{4}{|c|}{Stability} & \multicolumn{4}{|c|}{Secondary Structure}                                                                                                                                 \\ \hline
                        & \multicolumn{2}{|c|}{Full}         & \multicolumn{2}{|c|}{QLoRA}     & \multicolumn{2}{|c|}{Full}                & \multicolumn{2}{|c|}{QLoRA} & \multicolumn{2}{|c|}{Full} & \multicolumn{2}{|c|}{QLoRA}                                        \\ \hline
            Model       & BS                                 & LR                              & BS                                        & LR                          & BS                         & LR                          & BS & LR    & BS & LR    & BS & LR    \\ \hline
            ESM-2 8M    & 32                                 & 5E-05                           & 32                                        & 5E-05                       & 32                         & 5E-05                       & 32 & 5E-05 & 32 & 5E-05 & 32 & 5E-05 \\ \hline
            ESM-2 35M   & 32                                 & 5E-05                           & 32                                        & 5E-05                       & 32                         & 5E-05                       & 32 & 5E-05 & 32 & 5E-05 & 32 & 5E-05 \\ \hline
            ESM-2 150M  & 32                                 & 5E-05                           & 32                                        & 5E-05                       & 32                         & 5E-05                       & 32 & 5E-05 & 32 & 5E-06 & 32 & 5E-05 \\ \hline
            ESM-2 650M  & 16                                 & 5E-05                           & 16                                        & 5E-05                       & 16                         & 5E-05                       & 16 & 5E-05 & 16 & 5E-06 & 16 & 5E-05 \\ \hline
            ESM-2 3B    & 32                                 & 5E-06                           & 32                                        & 5E-05                       & 32                         & 5E-06                       & 32 & 5E-05 & 8  & 5E-06 & 4  & 5E-05 \\ \hline
            ESM C 300M  & 32                                 & 5E-05                           & 32                                        & 5E-05                       & 32                         & 5E-05                       & 32 & 5E-05 & 32 & 5E-05 & 32 & 5E-05 \\ \hline
            ESM C 600M  & 32                                 & 5E-05                           & 32                                        & 5E-05                       & 32                         & 5E-05                       & 32 & 5E-05 & 16 & 5E-06 & 32 & 5E-05 \\ \hline
            ProtBERT    & 32                                 & 5E-06                           & 32                                        & 5E-05                       & 32                         & 5E-06                       & 32 & 5E-05 & 32 & 5E-06 & 32 & 5E-05 \\ \hline
            ProtT5 Half & 8                                  & 5E-08                           & 8                                         & 5E-07                       & 32                         & 5E-06                       & 32 & 5E-05 & 4  & 5E-06 & 16 & 5E-05 \\ \hline
            Ankh Base   & 8                                  & 5E-05                           & 32                                        & 5E-06                       & 8                          & 5E-05                       & 8  & 5E-05 & 8  & 5E-05 & 8  & 5E-05 \\\hline
        \end{tabular}
\end{table}